\documentclass{article}

\makeatletter
\def\input@path{{./}}
\makeatother

\usepackage{preamble}
\DeclareMathOperator{\DecGap}{DecGap}
\DeclareMathOperator{\PriorGap}{PriorGap}
\usepackage{setspace}
\usepackage[
backend=biber,
style=alphabetic,
url=false,
isbn=false,
doi=false,
maxbibnames=10,
natbib=true,
backref=true]{biblatex}
\DefineBibliographyStrings{english}{%
  backrefpage = {cited on page},% originally "cited on page"
  backrefpages = {cited on pages},% originally "cited on pages"
}
\AtEveryBibitem{%
  \clearfield{volume}%
  \clearfield{number}
  \clearfield{pages}}
\DeclareFieldFormat{title}{\mkbibquote{#1}}
\title{Uncovering Latent Reasoning Strategies \\in Language Models}
\author{%
  Awni Altabaa \\
  Yale University \\
  \texttt{awni.altabaa@yale.edu} \\
  \And
  John Lafferty \\
  Yale University \\
  \texttt{john.lafferty@yale.edu} \\
}

\begin{document}

\maketitle

\begin{abstract}
A language model $p_\theta(y | x)$ trained on reasoning tasks learns to solve problems via multiple distinct strategies, yet these strategies are implicit and entangled within the model's response distribution.
We study the problem of decomposing the response distribution of a given pretrained language model into a structured, strategy-conditioned representation.
Specifically, we learn a latent-variable factorization $p_\theta(y| x) \leadsto (r_\phi(z | x),\, g_\phi(y | x, z))$, where a router $r$ maps each input to a distribution over latent strategies $z$ and a generator $g$ produces the response conditioned on that strategy. 
A key challenge is that the generator, initialized from the base model, already represents $p_\theta(y|x)$ without using $z$.
Standard variational inference therefore gives the model no incentive to route information through $z$ and can yield a severe form of posterior collapse.
To address this, we propose a variational objective that measures fractional information gain relative to the base model's response loss and concentrates reconstruction pressure on tokens with high base model surprisal, encouraging $z$ to encode strategy-relevant response variation.
We introduce a benchmark of multi-strategy algorithmic tasks and show that this objective recovers latent codes aligned with distinct reference strategies while preserving the base model's response distribution.\par\smallskip
\noindent\textbf{Code:}
\href{https://github.com/Awni00/latent-strategies-in-lms}%
     {\texttt{github.com/Awni00/latent-strategies-in-lms}}
%\begin{itemize}
%  \item A language model \(p_\theta(y|x)\) defines a distribution over response sequences. For reasoning tasks, this distribution is an entangled mixture over high-level strategies.
%  \item We formulate the problem of learning a latent-variable router-generator factorization \((r(z|x), g(y|x,z))\) of a \emph{pretrained} language model — recovering a disentangled \emph{probabilistic} representation of its implicit strategies.
%  \item We identify the challenges unique to this setting — most centrally that the pretrained generator already captures \(p(y|x)\) at initialization, causing a pathological form of posterior collapse that standard variational inference does not resolve.
%  \item We propose a variational training objective that turns absolute information gain into fractional information gain relative to the base model, while also placing targeted token-level pressure on \(z\) to encode information about \(y\) beyond what is predictable from \(x\).
%  \item We introduce a benchmark of multi-strategy algorithmic tasks and empirically evaluate our methodology on from-scratch models and pretrained language models.
%\end{itemize}
\end{abstract}

\section{Introduction}
\label{sec:introduction}

Language models trained on reasoning tasks can solve the same problem using multiple distinct strategies.
For example, a model asked to prove a theorem may produce a proof by contradiction, induction, or construction, while a model asked to solve a programming problem may formulate it as dynamic programming, graph search, or a greedy algorithm.
These strategies can recur across many problems, but are represented only implicitly.
The model's response distribution \(p_\theta(y|x)\) is an entangled mixture over distinct strategies, with no separate variable for identifying or controlling the strategy used during generation.

\begin{figure}[t]
  \centering
  \includegraphics[width=\textwidth]{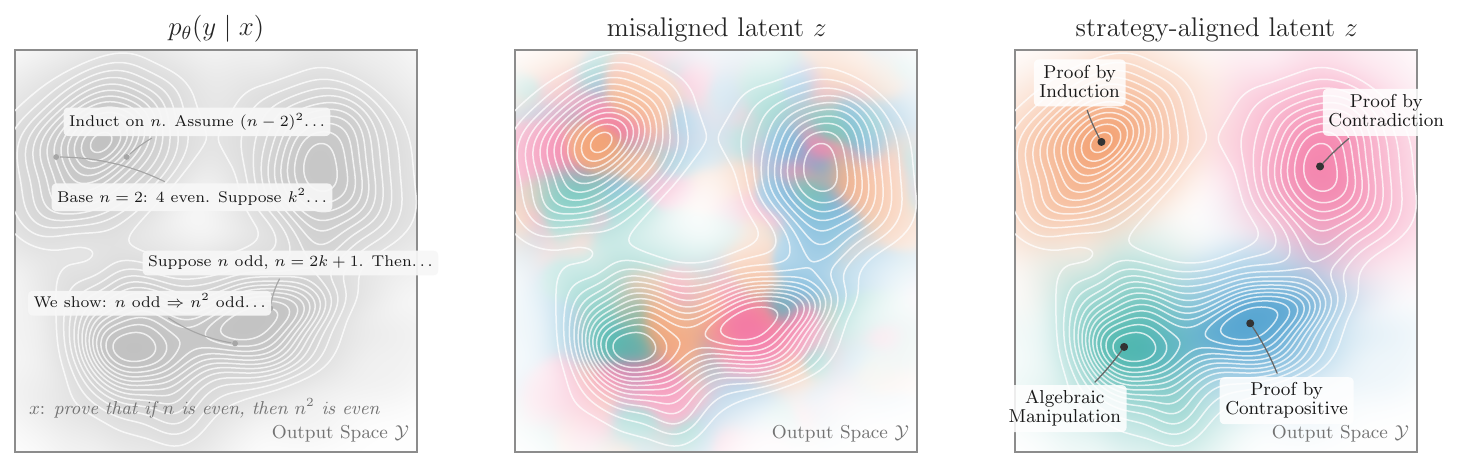}
  \caption{
\textbf{Latent strategy structure in output space.}
A pretrained LM's response distribution mixes strategies implicitly (left). A latent variable model initialized to match it leaves $z$ unused (center). Training aligns $z$ with reference strategies, disentangling them in latent space while preserving the base model's response distribution (right).
}
  \label{fig:motivation}
  \vskip-1.5em
\end{figure}

We study how to uncover these strategies by decomposing the response distribution of a pretrained language model into a structured, strategy-conditioned representation.
Concretely, we seek a \emph{router-generator factorization} $(r_\phi(z|x),\, g_\phi(y|x,z))$ of the base model \(p_\theta(y|x)\), where a router maps each input to a distribution over a latent strategy variable $z$, and a strategy-conditioned generator produces responses given the input and the sampled strategy.
The goal is to find a factorization that preserves the observable response distribution, so that $\int r_\phi(z|x) g_\phi(y|x,z) d z =: p_\phi(y|x) \approx p_\theta(y|x)$, while making $z$ informative and strategy-aligned: varying $z$ should change which strategy the generator uses, and the same latent value should carry the same strategy meaning across related inputs.

This problem is distinct from standard latent variable generative modeling \citep{kingma2013vae,sohn2015cvae}.
In VAE-style methods, the latent variable model is trained from data, and the target distribution is not already represented at initialization.
Here, the pretrained generator already captures $p_\theta(y|x)$ without any latent variable, creating a pathological form of posterior collapse.
A factorization that ignores $z$--by reproducing the base model's response distribution for every $z$--is available at initialization and is globally optimal for objectives that only fit the observable response distribution.
Standard variational training, including the conditional ELBO, is therefore subject to this failure mode \citep{bowman2016sentenceVae,he2019laggingInference,fu2019cyclicalAnnealing,hoffman2016elboSurgery}: it can preserve distributional fidelity while leaving $z$ inert, because it does not require $z$ to explain strategy-level variation.
% The problem is also distinct from mechanistic interpretability approaches that decompose internal representations post-hoc \citep{cunningham2023sparseAutoencoders,galichin2025reasoningSaes}, 
% because we seek a generative factorization that produces samples and supports intervention, not just a feature decomposition.
% \aawarning{Maybe defer any discussion of interpretability to the discussion or extended related work section; it's probably clear that these are distinct}

To address this problem, we propose a training objective that replaces marginal-fit pressure with \emph{model-directed reconstruction pressure}.
The fixed base model $p_\theta$ serves as a reference for what is already predictable from $x$ alone.
The reconstruction term is normalized by the base model's response loss and weighted at the token level by base model surprisal, concentrating pressure on positions where the base model leaves multiple continuations plausible, which tend to be the branch points where strategy choices are made.
Together, these mechanisms encourage \(z\) to encode strategy-relevant response variation that is not already determined by \(x\), rather than reconstructing what the base model already captures.

The broader promise of this problem is to expose latent capabilities that large language models already possess but do not reliably express under default sampling.
By recovering a strategy-aligned factorization, we can turn implicit strategy structure into an explicit object of analysis and intervention.
This has applications across model interpretability \citep{koh2020conceptBottleneck,cunningham2023sparseAutoencoders}, controllability \citep{keskar2019ctrl,li2021prefixTuning}, and exploration \citep{wang2023selfConsistency,yao2023treeOfThoughts}.
In particular, strategy-level interventions could support directed exploration across high-level solution strategies during reinforcement-learning post-training, beyond passive on-policy sampling from the base model's response distribution~\citep{yue2025rlvrBeyondBase,tuyls2025representationExploration}.
Such broader exploration could enable the discovery and improvement of previously under-sampled reasoning strategies and support their transfer or recombination across inputs.
We leave these downstream applications to future work.

We summarize the contributions of this paper as follows:

\begin{enumerate}[leftmargin=1em]
  \item \textbf{Problem formulation.} We formulate the problem of learning a router-generator factorization of a pretrained language model and identify the core challenge that makes this setting different from standard latent variable modeling.

  \item \textbf{Methodology.} We propose a variational objective with model-directed reconstruction pressure that uses the base model to normalize reconstruction in base model reference units and to direct token-level pressure toward positions of high base model residual surprisal, encouraging the latent to encode strategy-relevant variation.

  \item \textbf{Benchmark and empirical evaluation.} We introduce a controlled benchmark of multi-strategy algorithmic tasks with ground-truth strategy annotations, withheld during training and used only for evaluation.
  We evaluate our method on both pretrained and randomly initialized language models, characterizing recovery conditions, ablation effects, and failure modes.

  \item \textbf{Theoretical analysis.} We initiate a theoretical analysis of the strategy-structure recovery problem, studying the conditions under which a useful factorization can be identified.
\end{enumerate}

Our experiments show that strategy information is decodable from the base model's hidden states before any factorization is trained, indicating that the target structure exists.
Standard variational objectives nevertheless fail to expose this information through the latent variable.
Model-directed reconstruction pressure recovers latents that are aligned with reference solution strategies and semantically consistent across inputs.
As a result, the latent variable transfers the same strategy from one problem to another, while preserving the base model's response distribution.
This opens possibilities for intervening and more efficiently searching over solutions that are guided by high-level strategies.

\section{Problem Statement: Latent-Variable Factorization}
\label{sec:problem-statement}

% Establish that the target is a pretrained base model distribution to be factorized, not relearned.
Let \(p_\theta(y|x)\) denote the response distribution of a pretrained base model, where \(x\) is the input and \(y\) is the response sequence.
We consider reasoning tasks for which \(p_\theta(y|x)\) mixes multiple high-level strategies for producing valid responses.
The goal of this paper is to expose such variation as an explicit latent variable while preserving the observable response distribution that the base model already represents.
We formalize this goal as a \emph{latent-variable router-generator factorization} of \(p_\theta(y|x)\): the router \(r(z|x)\) maps each input to a distribution over latent strategy values \(z\), and the strategy-conditioned generator \(g(y|x,z)\) produces responses conditioned on both \(x\) and \(z\).
% Thus \(p_\theta(y|x)\) is not a target to be relearned from data; it is the already-represented response distribution whose latent strategy structure we aim to expose.

% Give the formal problem anchor: preserve the marginal while making the latent informative and strategy-aligned.
\begin{problemstatementbox*}[Latent-variable factorization of a pretrained model]\label{prob:factorization}
Let \(p_\theta(y|x)\) be the fixed response distribution of a pretrained base model.
A router-generator factorization \(p_\theta(y|x)\leadsto p_\phi=(r_\phi,g_\phi)\) introduces a latent variable \(z\in\mathcal Z\), a router \(r_\phi(z|x)\), and a strategy-conditioned generator \(g_\phi(y|x,z)\), inducing
\[
  p_\phi(y|x)=\int_{\mathcal Z} r_\phi(z|x)g_\phi(y|x,z)\,d\mu(z).
\]
The goal is to select a factorization that preserves the observable response distribution, \(p_\phi(y|x)\approx p_\theta(y|x)\), while making \(z\) informative about \(y\) beyond \(x\) and strategy-aligned: variation in \(z\) should expose high-level, reusable strategy variation in responses rather than arbitrary predictive surface details.
\end{problemstatementbox*}

% Explain why marginal fidelity alone leaves the desired factorization under-specified.
The objective cannot be merely to fit the latent marginal \(p_\phi(y|x)\) to the base model distribution.
For expressive router-generator classes, many factorizations can induce nearly the same distribution over \(y\) after integrating out \(z\).
One extreme ignores \(z\), with \(g_\phi(y|x,z)\) reproducing the base model distribution for all \(z\).
More generally, a marginal-preserving factorization may use \(z\) for input-specific partitions, surface form, or other predictive details rather than reusable strategy structure.
The problem is therefore not ordinary latent-variable modeling of \(p_\theta(y|x)\), but selection among marginal-preserving factorizations for one whose latent variable is informative and strategy-aligned.

% Explain requirements: informativeness, strategy-alignment, and cross-input semantic consistency.
The desired latent must first be informative: \(z\) should encode information about \(y\) beyond what is already determined by \(x\).
This is nontrivial for autoregressive language models, where the base model already defines a stochastic response distribution without an explicit latent; making \(z\) informative therefore requires it to encode structured variation across responses.
However, informativeness alone is too weak: a latent can help predict \(y\) by encoding surface form, input-specific partitions, or arbitrary details.
We seek an informative latent that is strategy-aligned, where a strategy is a high-level, reusable attribute of a response trace: the broad procedure used to generate a solution, not incidental wording, formatting, or token-level variation.
A strategy-aligned factorization is one in which changing \(z\) changes the generator along this strategy-level axis, while marginalizing over \(z\) remains faithful to the base model.

% Separate local strategy separation from the stronger cross-input consistency requirement.
There are two distinct requirements: local strategy separation and stable cross-input latent semantics.
Within-input strategy separation asks whether varying \(z\) changes the strategy used for a fixed input \(x\).
This rules out an inert latent, but not an input-local code whose interpretation changes across problems.
The stronger requirement is cross-input semantic consistency: a latent value or region should retain its strategy meaning across related inputs.
Only with this stability does \(z\) behave like a reusable strategy variable, rather than an input-specific response index.

% Bridge the abstract target to controlled evaluation and hand off to methodology.
To measure whether a factorization succeeds, we use controlled experiments in which response traces are parseable and their reference strategy can be classified.
These strategy labels are withheld during factorization training and used only to evaluate whether the learned latent achieves within-input strategy separation and cross-input semantic consistency.
The next section develops a procedure for learning a router-generator factorization whose latent variable is informative and strategy-aligned.

\section{Learning Strategy-Aligned Latent-Variable Factorizations}
\label{sec:methodology}

This section describes our methodology for learning a latent-variable router-generator factorization of a pretrained base model, with the goal of recovering an informative and strategy-aligned latent variable.
We first describe the architecture and parameterization of the router-generator model, then identify the under-specification and latent collapse issues that arise when applying standard variational inference to this setting, and finally introduce a modified training objective that addresses these issues by using the base model to direct reconstruction pressure.
Throughout, \(p_\theta(y|x)\) denotes the fixed pretrained base model whose response distribution we seek to factorize, \(\phi\) denotes the adapted factorized router-generator model's parameters, and \(\xi\) denotes the training-time posterior model parameters.

\subsection{A Transformer Architecture for Router-Generator Factorization}
\label{sec:methodology:parameterization}

\begin{figure}[t]
  \vskip-0.5em
  \centering
  \resizebox{\textwidth}{!}{%
    \input{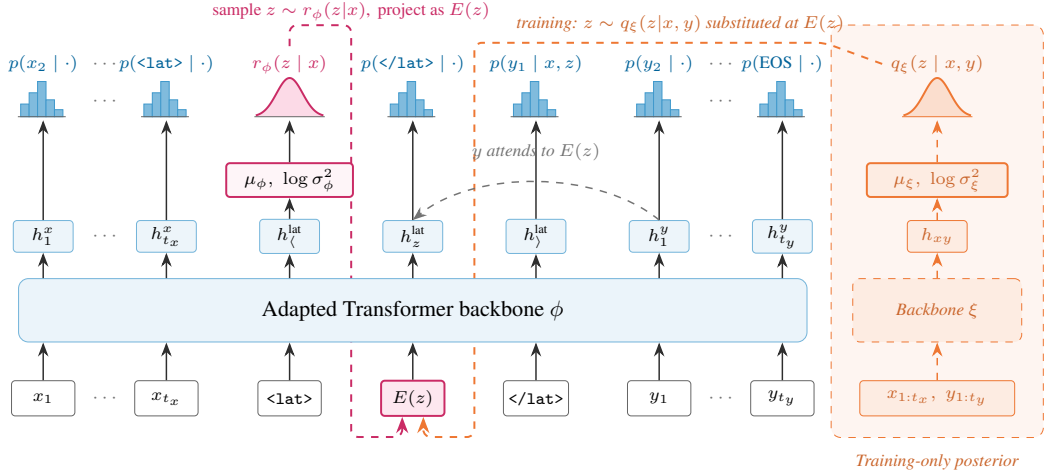}%
  }
  \vskip-0.5em
  \caption{%
    \textbf{Autoregressive router-generator Transformer architecture.}
    The adapted Transformer \(\phi\) implements both the router \(r_\phi(z|x)\), which reads the prompt and produces a latent distribution, and the strategy-conditioned generator \(g_\phi(y|x,z)\), which attends to \(E(z)\) while generating the response.
    During variational training, \(q_\xi(z|x,y)\) infers \(z\) from the full input-response pair.
  }
  \label{fig:architecture-ar-compact}
  \vskip -1.65em
\end{figure}

We implement the latent-variable router-generator factorization by \emph{adapting} a pretrained Transformer language model into the three roles shown in \Cref{fig:architecture-ar-compact}.
The router \(r_\phi(z|x)\) reads the input and produces a distribution over latent strategy values. Sampling from the router corresponds to selecting a strategy for generating a response.
The strategy-conditioned generator \(g_\phi(y|x,z)\) generates the response autoregressively while conditioning on \(x\) and an embedding \(E(z)\) of the latent strategy.
The training-time posterior \(q_\xi(z|x,y)\) reads the full input-response pair and produces a distribution over latent values, which is used to provide variational training signal.

This parameterization requires only small architectural additions to the base model.
Let \(h_{\phi}^{(L)}(x)\) denote the final-layer representation produced by the adapted Transformer after reading the input prefix \(x\).
The router adds two learned projection heads on top of this representation, producing the mean and log-variance of a diagonal Gaussian over the continuous latent:
\[
  r_\phi(z|x)
  =
  \mathcal N\!\left(
    W_\mu h_{\phi}^{(L)}(x),
    \mathrm{diag}\!\left(\exp(W_\sigma h_{\phi}^{(L)}(x))\right)
  \right).
\]
During autoregressive generation, a latent sample \(z\) is projected to an embedding \(E_\phi(z)\) and inserted as a pseudo-token at the embedding layer, so response tokens can attend to it while the generator predicts \(y\).
The training-time posterior \(q_\xi(z|x,y)\) is implemented similarly to the router, parameterizing a Gaussian distribution over \(z\) from the final-layer input representation after reading \((x,y)\).
%  the full input-response pair. 
% We write \(p_\phi(y|x):= \int_{\mathcal Z} r_\phi(z|x) g_\phi(y | x,z) \,d\mu(z)\) for the marginal distribution induced by the router-generator factorization.
% The training-time posterior is implemented analogously, using a separate adapted Transformer to read the full input-response sequence and parameterize
% \[
%   q_\xi(z|x,y)
%   =
%   \mathcal N\!\left(
%     U_\mu h_{\xi}^{(L)}(x,y),
%     \mathrm{diag}\!\left(\exp(U_\sigma h_{\xi}^{(L)}(x,y))\right)
%   \right).
% \]

The router and generator are a single autoregressive model with shared parameters \(\phi\).
The posterior is a separate model with parameters \(\xi\), and is used only during variational training.
Both models are initialized from the base parameters \(\theta\) and trained as light adaptations using LoRA-style low-rank updates~\citep{hu2022lora}, together with the small Gaussian heads and latent embedding projection.

% this below takes up space and is unnecessary
% At sampling time, we draw \(z\sim r_\phi(\cdot|x)\) and then draw \(y\sim g_\phi(\cdot|x,z)\).
% The induced latent marginal is
% \[
%   p_\phi(y|x)
%   =
%   \int_{\mathcal Z}
%   r_\phi(z|x)g_\phi(y|x,z)\,d\mu(z).
% \]
% This notation emphasizes that \(p_\phi(y|x)\) is the marginal distribution after integrating out \(z\), not a z-free pass through the adapted generator.

Because \(p_\theta(y|x)\) is already represented by the base model, lightweight adaptation lets the generator inherit the base model's response distribution rather than relearn it.
However, because the base generator realizes the target response distribution, a factorization that ignores \(z\) is available at initialization, biasing optimization toward it when the objective only measures marginal fit.
The training objective must therefore choose a factorization where \(z\) explains meaningful strategy variation.

\subsection{ELBO Under-Specification and Latent Collapse}
\label{sec:methodology:elbo-fails}

The standard negative conditional ELBO objective for this router-generator model is
% \begin{equationbox}
\begin{equation}\label{eq:conditional-elbo}
  \calJ_{\mathrm{cELBO}}(x,y;\phi,\xi)
  =
  \expectunder{z\sim q_\xi(\cdot|x,y)}{
    -\log g_\phi(y|x,z)
  }
  +
  \mathrm{KL}\!\left(
    q_\xi(\cdot|x,y)
    \,\Vert\,
    r_\phi(\cdot|x)
  \right).
\end{equation}
% \end{equationbox}
Let \(p_\phi(z|x,y) = r_\phi(z|x) g_\phi(y | x,z) / p_\phi(y|x)\) denote the posterior induced by the router-generator factorization. This objective satisfies the identity
% \begin{equationbox}
\begin{equation}\label{eq:conditional-elbo-identity}
  \calJ_{\mathrm{cELBO}}(x,y;\phi,\xi)
  =
  -\log p_\phi(y|x)
  +
  \mathrm{KL}\!\left(
    q_\xi(\cdot|x,y)
    \,\Vert\,
    p_\phi(\cdot|x,y)
  \right).
\end{equation}
% \end{equationbox}
Thus the standard variational objective asks the induced marginal \(p_\phi(y|x) = \int_{\mathcal Z} r_\phi(z|x) g_\phi(y | x,z) \,d\mu(z)\) to match the observed response distribution, and asks \(q_\xi\) to match the posterior of whichever factorization produced that marginal.
It does not ask whether \(z\) explains variation in \(y\) beyond \(x\), nor whether the resulting latent is strategy-aligned.

This is an objective-level under-specification problem.
If the generator class can represent the base model while ignoring \(z\), then the factorization
\[
  g_\phi(y|x,z)=p_\theta(y|x)
  \quad\text{for all }z,
  \qquad
  q_\xi(z|x,y)=r_\phi(z|x)
\]
\begin{wrapfigure}{r}{0.45\textwidth}
  \vskip -1.25em
  \centering
  \input{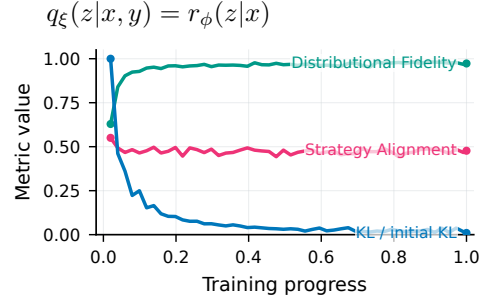}
  \vskip -1.5em
  \caption{\textbf{ELBO training can preserve fidelity while leaving the latent unused.}
  Distributional fidelity rises, posterior-router KL collapses relative to initialization, and strategy alignment remains low.
  }
  \label{fig:elbo-fails}
  \vskip -3em
\end{wrapfigure}
preserves the base model's response distribution and attains the objective's optimum.
The latent variable is then inert even though \(p_\phi(y|x)=p_\theta(y|x)\). 
In our setting, with expressive generator classes built from pretrained language models, this \(z\)-ignoring factorization is not only available, but also close to initialization, biasing optimization toward it.
In particular, the generator can implement \(g_\phi(y|x,z) = p_\theta(y|x)\) for all \(z\) by simply learning to ignore \(z\). 
This is a latent collapse driven by both the objective's under-specification and the proximity of the inert solution to initialization.

\subsection{Model-Directed Reconstruction for Informative, Strategy-Aligned Latents}
\label{sec:methodology:excess-reconstruction}

The ELBO failure above shows that marginal fidelity is not enough: a factorized model can preserve \(p_\theta(y|x)\) while leaving \(z\) unused.
We address this selection problem with a \emph{model-directed reconstruction term}.
The base model \(p_\theta\) is used to concentrate reconstruction pressure on response tokens where the base model is uncertain, so the latent is encouraged to explain variation in \(y\) that the base model does not already explain through \(x\) and the autoregressive prefix.

The fixed base model is used first to define the response distribution being factorized.
Training examples are drawn as
\[
    x\sim\mathcal D_X,
    \qquad
    y\sim p_\theta(\cdot|x), \qquad
    z\sim q_\xi(\cdot|x,y).
\]
Thus \(y\) is sampled from the base model, while \(z\) is inferred by the training-time posterior for that sampled input-response pair.
No strategy annotations are available during training.

The core methodological idea is to use the base model \(p_\theta\) to \emph{adaptively} rescale the reconstruction loss and direct reconstruction pressure towards tokens where the base model is uncertain.
We instantiate this idea with token-level base model surprisal. 
Define
\[
  b_{\theta,t}(x,y)
  =
  -\log p_\theta(y_t|x,y_{<t}), \qquad
  b_\theta(x,y)
  =
  \frac{1}{T_y}
  \sum_{t=1}^{T_y}
  b_{\theta,t}(x,y),
\]
as the $z$-free base model surprisal.
This score is computed by the latent-free base model.
\(b_{\theta,t}(x,y)\) measures how much uncertainty remains about \(y_t\) after conditioning on \(x\) and \(y_{<t}\), and therefore gives a local estimate of where \(z\) can explain variation beyond the $z$-free prefix model.

This token-level view matters because autoregressive responses quickly become partially self-predictable.
After an early commitment, many later tokens may be determined by \(x\) and \(y_{<t}\), even if the commitment itself reflects a strategy choice.
Our reconstruction term therefore replaces uniform token pressure with \(p_\theta\)-surprisal-directed pressure, assigning more weight to branch points and high-surprisal positions where the base model leaves multiple continuations plausible.

\begin{figure}[t]
  \centering
  \input{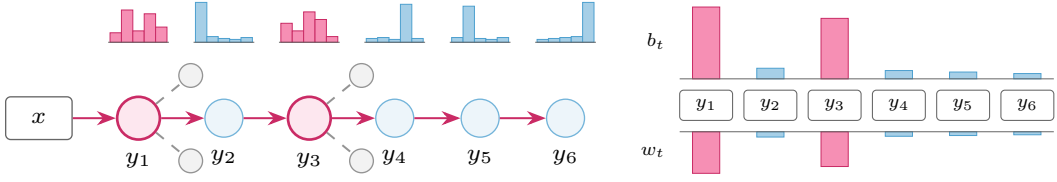}
  \caption{%
    \textbf{Token-level \(p_\theta\) surprisal weighting.}
    The left panel shows a response path with branch points, where several next tokens remain plausible, and determined positions, where the next token is mostly fixed by \(x\) and earlier response tokens.
    The right panel shows z-free base model surprisal \(b_{\theta,t}(x,y)\) above each token and the corresponding token weight below it; higher-surprisal positions receive more reconstruction pressure in \(a_{\theta,t}(x,y)\).%
  }
  \label{fig:token-weighting}
  \vskip -1em
\end{figure}

The proposed objective is
\begin{equationbox}
  \begin{equation}\label{eq:reconstruction-objective}
    \begin{aligned}
      \calJ_\theta(x,y;\phi,\xi)
      &=
      \calR_\theta(x,y;\phi,\xi)
      +
      \beta\,
      \mathrm{KL}\!\left(
        q_\xi(\cdot|x,y)
        \,\Vert\,
        r_\phi(\cdot|x)
      \right), \\
      \calR_\theta(x,y;\phi,\xi)
      &=
      \frac{1}{c_\theta}
      \expectunder{z\sim q_\xi(\cdot|x,y)}{
        \sum_{t=1}^{T_y}
        a_{\theta,t}(x,y)\,
        \bigl[-\log g_\phi(y_t|x,z,y_{<t})\bigr]
      }, \\
      a_{\theta,t}(x,y)
      &=
      \alpha \, \frac{1}{T_y}
      +
      (1-\alpha)\, w_{\theta,t}^{(\gamma)}(x,y),
      \qquad
      c_\theta
      =
      \expectunder{x\sim\mathcal D_X,\;y\sim p_\theta(\cdot|x)}{
        b_\theta(x,y)
      }.
    \end{aligned}
  \end{equation}
\end{equationbox}
Quantities with a \(\theta\) subscript reflect values that depend on the base model $p_\theta$.
The global scale \(c_\theta\) expresses reconstruction in base model reference units, so the reconstruction term is normalized across different data distributions and remains comparable to the KL term even when the base model already has low $z$-free loss.
The interpolation parameter \(\alpha\in[0,1]\) mixes uniform response-token pressure with surprisal-directed pressure. The token weights $w_{\theta,t}^{(\gamma)}(x,y)$ are defined in terms of the base model surprisal as follows:
\begin{equationbox}
  \begin{equation}\label{eq:token-reconstruction-weights}
    \widetilde w_{\theta,t}^{(\gamma)}(x,y)
    =
    \frac{
      b_{\theta,t}(x,y)^\gamma
    }{
      \sum_{s=1}^{T_y} b_{\theta,s}(x,y)^\gamma
    }, \qquad
    w_{\theta,t}^{(\gamma)}(x,y)
    =
    \frac{
      \widetilde w_{\theta,t}^{(\gamma)}(x,y)
    }{
      \kappa_\theta^{(\gamma)}(x,y)
    },
  \end{equation}
\end{equationbox}
where 
\(
\kappa_\theta^{(\gamma)}(x,y) =
\frac{1}{b_\theta(x,y)} \,
\sum_{s=1}^{T_y}
\widetilde w_{\theta,s}^{(\gamma)}(x,y)
\, b_{\theta,s}(x,y)
\)
is a normalization factor that preserves the $z$-free base scale of the reconstruction term. The parameter \(\gamma\ge0\) controls concentration of the token-level weighting. The reconstruction term \(\calR_\theta\) is designed so that \(\bbE_{x,y}\bra{\calR_\theta(x,y;\phi,\xi)} = 1\) when \(g_\phi(y|x,z,y_{<t}) = p_\theta(y_t|x,y_{<t})\), and thus can be interpreted as a (weighted) fraction of the base model response loss that remains after conditioning on the latent \(z\).
The objective therefore balances the KL penalty against the \emph{fraction} of the base model's response loss explained by \(z\), giving reconstruction improvement a fractional information-gain interpretation rather than measuring it in absolute log-loss units.

\section{Experiments}
\label{sec:experiments}

\subsection{Benchmark and evaluation protocol}
\label{sec:experiments:setup}

We evaluate on a controlled multi-strategy algorithmic benchmark designed to make strategy alignment measurable.
The task structure is parseable, allowing generated solution traces to be validated and matched to compatible benchmark strategies.
This enables ground-truth evaluation of whether a learned latent variable tracks strategy-level variation, a property that is not directly observable in open-ended language modeling.

Each example consists of a problem instance \(x\) and a solution trace \(y\).
The benchmark covers six task families: list summation, sorting algorithms, grid pathfinding, linear equation solving, base conversion, and multidigit addition.
Solution traces instantiate task-specific procedures for reaching the same valid answer, such as different traversal orders, algebraic manipulation orders, or algorithms.
Details of the benchmark design are provided in~\Cref{app:benchmark-data-generation}.

We use a two-stage experimental protocol. 
First, to construct a controlled factorization target with known reference strategies, we fit a language model to the benchmark distribution, producing a base model \(p_\theta(y|x)\) whose response distribution mixes multiple strategies.
We then train a router-generator factorization on samples from this fixed response distribution,
\[
  (x,y)\sim\mathcal D
  \quad\text{for base model fitting},
  \qquad
  x\sim\mathcal D_X,\;
  y\sim p_\theta(\cdot|x)
  \quad\text{for factorization training}.
\]

We report two primary evaluation axes. 
Distributional Fidelity measures whether the factorized model preserves the observable base model behavior, in particular whether it continues to produce valid solution traces. 
Strategy Alignment measures whether the learned latent is aligned with the reference strategy structure, in particular whether it achieves cross-input semantic consistency in assigning the same latent value to the same strategy across different problems. 
For each method class, we report the best run over the swept KL-to-reconstruction weight \(\beta\); sweep details are in~\Cref{app:experimental-details}.

\subsection{Recovering Strategy-Aligned Latents}
\label{sec:experiments:strategy-aligned-latents}

\paragraph{Implicit strategy information.}
We first ask whether the base model already contains information about the reference strategy before any router-generator factorization is trained.
For each task and token position, we train a lightweight linear probe on the base model's hidden states to predict the reference strategy annotation.

\begin{wrapfigure}{r}{0.50\textwidth}
  \vspace{-2.5ex}
  \centering
  \includegraphics[width=\linewidth]{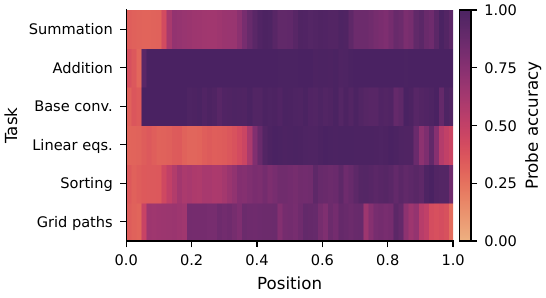}
  \vskip-1ex
  \caption{\textbf{Strategy labels are decodable from base model hidden states.}
  Rows are benchmark tasks, columns are normalized token positions, and color gives linear-probe accuracy for predicting the ground-truth strategy from the base model's hidden states.}
  \label{fig:entanglement}
  \vspace{-4ex}
\end{wrapfigure}
\Cref{fig:entanglement} shows that strategy information is broadly decodable across the algorithmic benchmark, especially at positions where strategy-specific computation is expressed in the sequence.
This provides evidence that the target strategy structure is represented in the base model's hidden states, but it is not exposed as a reusable latent variable that can be sampled, transferred, or compared across inputs.

% \begin{findingbox}[The base model represents implicit strategy information]
% \label{find:implicit-strategy-structure}
% Strategy information is decodable from the base model's hidden states before factorization.
% This suggests that the target structure is present in the model but not exposed as a usable latent variable.
% \end{findingbox}

\begin{figure}[t]
  \centering
  \includegraphics[width=\textwidth]{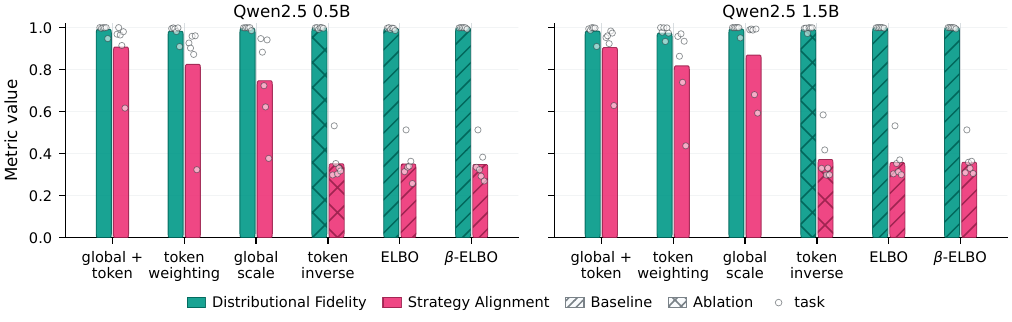}
  \caption{\textbf{Method comparison on pretrained Qwen2.5 runs.}
  For each objective variant and model size, bars show Distributional Fidelity and Strategy Alignment, with open points giving per-task values.
  Strategy Alignment denotes cross-input latent strategy alignment measured by Analogical Consistency.
  Hatching marks baselines and ablations.}
  \label{fig:reconstruction-decodability}
\end{figure}

\paragraph{ELBO failure.}
The presence of implicit strategy information does not imply that a standard variational objective will identify it.
The conditional ELBO is under-specified for this purpose: it rewards matching the observable response distribution, but does not require \(z\) to explain strategy-level variation.
Empirically, \Cref{fig:elbo-fails} and \Cref{fig:reconstruction-decodability} show that ELBO-style baselines can preserve Distributional Fidelity while leaving the latent inert or strategy-unaligned.

% \begin{findingbox}[ELBO training preserves behavior without identifying strategy]
% \label{find:elbo-under-specified}
% Standard variational objectives can maintain Distributional Fidelity while leaving \(z\) inert or unaligned with reference solution strategies.
% This shows that fitting the observable response distribution is not sufficient to select a strategy-aligned factorization.
% \end{findingbox}

\begin{figure}[t]
  \centering
  \includegraphics[width=\textwidth]{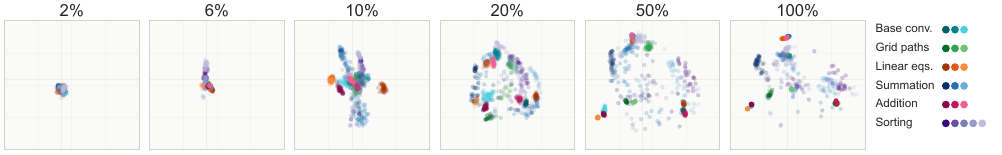}
  \caption{\textbf{Posterior latent geometry over training.}
  Each panel is a PCA projection of posterior mean latents at the indicated fraction of training.
  Points are colored by task and strategy; the initially overlapping cloud separates into task- and strategy-specific regions over training.}
  \label{fig:disentanglement}
  \vskip -1.75ex
\end{figure}

The proposed model-directed reconstruction objective changes which marginal-preserving factorization is learned.
\Cref{fig:reconstruction-decodability} depicts the distributional fidelity and strategy alignment of factorizations of pretrained language models induced by different objectives.
The model-directed reconstruction objective and its variants recover strategy-aligned latents while preserving the base model's response distribution, whereas ELBO-style baselines preserve the distribution but fail to recover strategy-aligned latents.
The latent dynamics in \Cref{fig:disentanglement} provide complementary geometric evidence: posterior latents that begin as an overlapping cloud separate into task- and strategy-structured regions over training.
In the multi-task setting, this organization is task-conditioned rather than a single global strategy codebook: the same latent regions can carry different strategy meanings for different task families.
This latent multiplexing is visible in linear diagnostics, as strategies pooled across all tasks are not linearly separable from \(z\) alone, but become linearly separable when task identity is provided.
Together, these results indicate that model-directed reconstruction pressure can expose the strategy structure that was implicit in the base model's response distribution.

% \begin{findingbox}[Model-directed reconstruction recovers strategy-aligned latents]
% \label{find:strategy-aligned-factorization}
% The proposed \(p_\theta\)-directed reconstruction objective preserves distributional fidelity while recovering latent structure aligned with reference solution strategies.
% This alignment is not achieved by standard variational objective baselines.
% \end{findingbox}

\paragraph{Cross-input semantic consistency.}
A latent variable can separate strategies locally for each input while failing to give \(Z\) a stable strategy meaning across inputs.
We therefore evaluate a stronger notion of Strategy Alignment with Analogical Consistency: sample a source input \(X\), draw \(Z\sim r_\phi(\cdot|X)\), reuse the same \(Z\) on a related target input \(X'\), and check whether \(Y\sim g_\phi(\cdot|X,Z)\) and \(Y'\sim g_\phi(\cdot|X',Z)\) use the same inferred strategy.
In the main figures, including \Cref{fig:reconstruction-decodability}, ``Strategy Alignment'' denotes this stronger cross-input metric, which we write as \(\operatorname{AnalogicalConsistency}=\Pr[\operatorname{strat}(Y)=\operatorname{strat}(Y')]\).
% \Cref{fig:analogical-consistency} is a metric-definition schematic; the quantitative evidence for this property is the Strategy Alignment metric in \Cref{fig:reconstruction-decodability}.
High Strategy Alignment therefore means that \(Z\) carries the same strategy meaning across inputs; \Cref{fig:reconstruction-decodability} shows that our objective recovers this property.

\paragraph{Ablation: \emph{inverse}-surprisal reconstruction pressure.}
The proposed objective puts more reconstruction pressure on high-surprisal tokens under \(p_\theta\), where strategy-branching information is expected to concentrate.
We test this hypothesis through an ablation in which the reconstruction loss is weighted by \emph{inverse} surprisal, so that low-surprisal tokens receive more emphasis.
This ablation weakens Strategy Alignment, supporting the view that the model-directed reconstruction helps by pressuring \(z\) to encode strategy-relevant response variation.
Details are in~\Cref{app:experimental-details}.

% \begin{wrapfigure}{r}{0.52\textwidth}
% %   \vspace{-2ex}
%   \centering
%   \includegraphics[width=\linewidth]{../shared/figures/analogical-consistency/figure.pdf}
% %   \vskip-1ex
%   \caption{\textbf{Analogical Consistency evaluation.}
%   The schematic defines the intervention used to measure cross-input semantic consistency: a latent \(z\) inferred from source input \(X\) is reused when generating from target input \(X'\), and the generated outputs \(Y\) and \(Y'\) are compared for whether they use the same strategy.}
%   \label{fig:analogical-consistency}
% %   \vspace{-1.5ex}
% \end{wrapfigure}

% \begin{findingbox}[Recovered latents are semantically consistent across inputs]
% \label{find:cross-input-semantic-consistency}
% Our method recovers latents with cross-input semantic consistency, not merely within-input strategy separation.
% Analogical Consistency measures this property by testing whether reusing the same \(z\) transfers the generated solution strategy from \(X\) to \(X'\).
% \end{findingbox}

\subsection{How Model-Directed Reconstruction Pressure Drives Strategy Alignment}
\label{sec:experiments:alignment-reconstruction}

The preceding results show that the proposed objective recovers strategy-aligned latents.
We next ask which part of the training signal explains the alignment gains across proposed methods and variational baselines.
If alignment comes from capturing high-surprisal response variation in \(z\), runs with lower token-weighted reconstruction loss should have higher Strategy Alignment.

\begin{figure}[t]
  \centering
  \begin{minipage}[t]{0.49\textwidth}
    \vspace{0pt}
    \centering
    \includegraphics[height=0.716\linewidth]{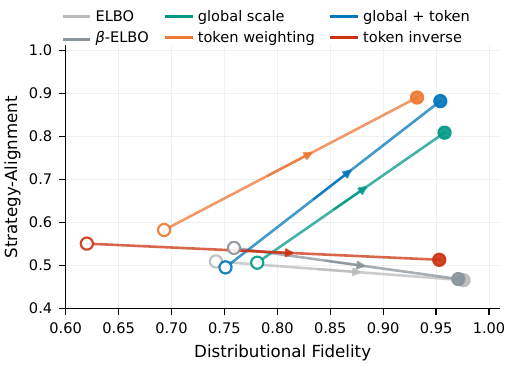}
  \end{minipage}\hfill
  \begin{minipage}[t]{0.49\textwidth}
    \vspace{0pt}
    \centering
    \includegraphics[height=0.716\linewidth]{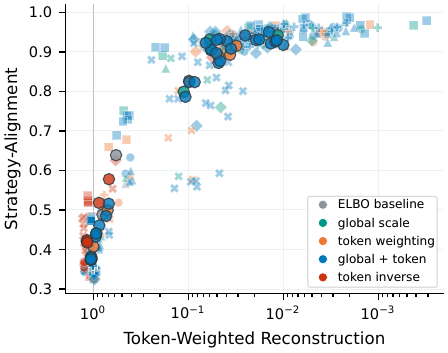}
  \end{minipage}
  \caption{
    \textbf{Left:} training trajectories in fidelity-alignment phase space for different objective variants.
    \textbf{Right:} token-weighted reconstruction loss versus Strategy Alignment at training endpoints; outlined points show method centroids and translucent points show individual runs.
  }
  \label{fig:alignment-diagnostics}
  \label{fig:fidelity-alignment-phase-space}
  \label{fig:token-reconstruction-strategy-alignment-scatter}
  \vskip -1.5em
\end{figure}

\Cref{fig:alignment-diagnostics} supports this mechanism-level interpretation.
The phase-space dynamics show that, when the objective is underspecified, the model can move toward high Distributional Fidelity without moving toward high Strategy Alignment.
The scatter plot shows that token-weighted reconstruction loss tracks Strategy Alignment across all methods.
This suggests strategy alignment improves when \(z\) is pressured to explain strategy-relevant residual surprisal under \(p_\theta\).

% \begin{findingbox}[Model-directed reconstruction pressure explains alignment gains]
% \label{find:alignment-reconstruction-frontier}
% Token-weighted reconstruction loss tracks strategy alignment across both proposed methods and variational baselines.
% This supports the mechanism that alignment improves when \(z\) is pressured to explain response regions where the base model has strategy-relevant residual surprisal.
% \end{findingbox}

% Standard ELBO-style objectives can fit observable behavior without moving \(z\) into this explanatory role.
% By contrast, the \(p_\theta\)-directed objective uses the base model to emphasize residual uncertainty, making the latent variable account for response variation that is not determined by \(x\) alone and relevant to strategy choice.
% The next section studies this selection problem in a simplified theoretical model.

\section{Toward a Theory of Strategy-Structure Recovery}
\label{sec:theory}

This section outlines a line of analysis that can provide theoretical insight into the methods introduced in the paper.
The goal is to understand when the variational approach can identify a latent variable \(Z\) that aligns with a target strategy structure \(S^\star\) implicit in the pretrained model.
We provide a preview of the main results here; the full formal treatment is given in \Cref{app:variational-complexity-theory}.

Consider finite sets \(\calX\) and \(\calY\), an input law \(\mu\), and a fixed pretrained response distribution \(p(y\mid x)\).
For an encoder \(q(z\mid x,y)\), router \(r(z\mid x)\), and generator \(g(y\mid x,z)\), let \(Q_q\) be the joint law obtained by sampling \(X\sim\mu\), \(Y\sim p(\cdot\mid X)\), and \(Z\sim q(\cdot\mid X,Y)\).
We study, for \(\beta\in[0,1)\) and \(\lambda\ge0\), the population complexity-penalized variational objective
\begin{equation}
\begin{aligned}
  \calL_{\beta,\lambda}(q,r,g) &:={} \Expect_{Q_q}[-\log g(Y\mid X,Z)] + \beta\,\Expect_{X,Y} \mathrm{KL}(q(\cdot\mid X,Y)\|r(\cdot\mid X)) + \lambda C(r,g),
\end{aligned}
\end{equation}
where \(C(r,g)\) is a theoretical proxy for architecture, parameterization, and optimization biases that favor simple router-generator factorizations.
The appendix shows the decomposition
\begin{equation}
\begin{aligned}
  \calL_{\beta,\lambda}(q,r,g) &={} H(Y\mid X) -(1-\beta)I_q(Y;Z\mid X) \\ &\quad+ \DecGap(q,g) + \beta\PriorGap(q,r) + \lambda C(r,g),
\end{aligned}
\end{equation}
where \(\DecGap\) and \(\PriorGap\) are the generator and router KL gaps.
% Thus \(\beta<1\) supplies information pressure, while the router-generator terms select which informative latent variables can be realized simply and with high fidelity.
For a fixed encoder \(q\), define
\begin{equation}
  A_{\beta,\lambda}(q) := \inf_{r\in\calR,\;g\in\calG} \left[ \DecGap(q,g)+\beta\PriorGap(q,r)+\lambda C(r,g) \right],
\end{equation}
We show that minimizing \(\calL_{\beta,\lambda}\) is equivalent to maximizing the following profiled score
\begin{equation}
  \calJ^{\mathrm{prof}}_{\beta,\lambda}(q) := I_q(Y;Z\mid X)-\frac{A_{\beta,\lambda}(q)}{1-\beta}.
\end{equation}
% Since \(\inf_{r,g}\calL_{\beta,\lambda}(q,r,g)=H(Y\mid X)-(1-\beta)\calJ^{\mathrm{prof}}_{\beta,\lambda}(q)\), profiling over \(r,g\) selects encoders by maximizing \(\calJ^{\mathrm{prof}}_{\beta,\lambda}\).
The first term \(I_q(Y;Z\mid X)\) is the information pressure that encourages \(Z\) to capture information about \(Y\) beyond what is already predictable from \(X\), while the second term \(A_{\beta,\lambda}(q)\) is an effective approximation-complexity penalty that measures whether the induced distribution is close to the target while being simple, as measured by the complexity term \(C(r,g)\).

Let \(S^\star=f(X,Y)\) be the targeted strategy, and let \(q^\star(z | x,y)= \Ind{z=S^\star}\).
Let \(d(q,S^\star)\) be a recovery discrepancy between \(q\) and \(S^\star\).
% ; the appendix instantiates \(d\) with information-theoretic distances capturing weak and strong recovery.
Our primary result gives sufficient conditions under which the complexity-penalized variational objective selects a latent factorization aligned with \(S^\star\).
% , together with sample-complexity conditions for empirical recovery.

\begin{featurebox}
\textbf{Theoretical analysis of latent-variable disentanglement}
\par
Given a discrepancy measure \(d(q,S^\star)\), define
\begin{equation}
  \Delta_{\beta,\lambda}^{\mathrm{prof}}(\delta) = \calJ^{\mathrm{prof}}_{\beta,\lambda}(q^\star)\; - \!\sup_{q\in\calQ:\; d(q,S^\star)\ge \delta} \calJ^{\mathrm{prof}}_{\beta,\lambda}(q).
\end{equation}
If \(\Delta_{\beta,\lambda}^{\mathrm{prof}}(\delta)>0\), then every population minimizer
\(
  (\widehat q,\widehat r,\widehat g) \in \argmin_{q, r, g} \calL_{\beta,\lambda}(q,r,g)
\)
satisfies \(d(\widehat q,S^\star)<\delta\).
Moreover, if the empirical objective \(\widehat{\calL}_{\beta,\lambda,n}\) satisfies, with probability at least \(1-\eta\),
\[
  \sup_{q,r,g} \left| \widehat{\calL}_{\beta,\lambda,n}(q,r,g) - \calL_{\beta,\lambda}(q,r,g) \right| \le \frac{(1-\beta)\Delta_{\beta,\lambda}^{\mathrm{prof}}(\delta)}{4},
\]
then every empirical minimizer \((\widehat q_n,\widehat r_n,\widehat g_n)\in\argmin_{q,r,g}\widehat{\calL}_{\beta,\lambda,n}(q,r,g)\) satisfies \(d(\widehat q_n,S^\star)<\delta\).
For finite \(\calQ,\calR,\calG\) with sub-Gaussian losses,
\[
  n = \calO\!\left(
  \frac{\log(|\calQ||\calR||\calG|)+\log(1/\eta)}
  {(1-\beta)^2 (\Delta_{\beta,\lambda}^{\mathrm{prof}}(\delta))^2} \right)
\]
is sufficient to ensure this condition.
\vskip5pt
\end{featurebox}
% The margin can be positive or negative; if it is negative, the objective need not align \(Z\) with \(S^\star\).
% Positive margin requires the target strategy to be predictive and factorization-compatible, and the profiled router-generator complexity is the inductive bias that can make it beat competing high-information attributes.

\section{Conclusion}
\label{sec:discussion}

This paper studies the problem of recovering explicit latent variable representations of reasoning strategies from pretrained language models.
The central observation is that a pretrained model's response distribution is an entangled mixture over high-level strategies, and that exposing this structure requires more than fitting a latent variable model to observable outputs using previously explored methods.
In particular, because the base generator already realizes this response distribution, a factorization that ignores the latent variable is immediately available and is globally optimal under objectives that only measure marginal fit.

We show that standard variational training can preserve the observable response distribution while leaving the latent variable unused.
To address this, we propose a model-directed reconstruction objective that uses the base model itself as a reference.
In our approach, reconstruction is normalized by the base model's response loss and weighted by the uncertainty seen in the base model at the token level, allowing the loss to concentrate on positions where strategy choices are most consequential.
On a controlled benchmark of multi-strategy algorithmic tasks, this objective recovers latent structure that is aligned with reference solution strategies and is semantically consistent across inputs. 
% In this way, the latent variable transfers the same strategy from one problem to another.

The benchmark design makes strategy recovery measurable, but it limits the scope of the conclusions.
The benchmarks we introduce cover synthetic algorithmic tasks with ground-truth strategies; how well the methodology extends to open-ended reasoning tasks remains an open question.
Moreover, the theoretical analysis is preliminary, and the connection between the oracle selection criterion and actual training dynamics is not yet formally characterized. 
Further discussion of related work, broader implications, and future directions appears in \Cref{app:extended-discussion}.

\printbibliography
\clearpage
\appendix
\section{Extended Discussion}
\label{app:extended-discussion}

This paper studies the problem of recovering explicit latent variable representations of reasoning strategies from pretrained language models. The central observation is that a pretrained model's response distribution is an entangled mixture over high-level strategies, and that exposing this structure requires more than fitting a latent variable model to observable outputs using previously explored methods.
In particular, because the base generator already realizes this response distribution, a factorization that ignores the latent variable is immediately available and is globally optimal under objectives that only measure marginal fit.

We show that standard variational training can preserve the observable response distribution while leaving the latent variable unused. To address this, we propose a model-directed reconstruction objective that uses the base model itself as a reference. In our approach, reconstruction is normalized by the base model's response loss and weighted by the uncertainty seen in the base model at the token level, allowing the loss to concentrate on positions where strategy choices are most consequential. On a controlled benchmark of multi-strategy algorithmic tasks, this objective recovers latent structure that is well-aligned with reference solution strategies, while also being semantically consistent across inputs. In this way, the latent variable transfers the same strategy from one problem to another.

\subsection{Related Work}

There is a rich and extensive literature on latent variable methods for generative models of text. This work includes VAEs~\citep{kingma2013vae} and CVAEs~\citep{sohn2015cvae}, text and sequence VAEs~\citep{bowman2016sentenceVae,zhao2017dialogCvae}, and disentanglement-oriented variants such as the $\beta$-VAE~\citep{higgins2017betavae,burgess2018betavae}, FactorVAE~\citep{kim2018factorvae}, and InfoGAN~\citep{chen2016infogan}, as well as pretrained-Transformer VAEs such as OPTIMUS~\citep{li2020optimus} and the Transformer VAE of~\citet{park2021transformerVae}. Adjacent language-VAE work also uses structured latent spaces to separate syntax from semantics, represent discrete linguistic factors, and control semantic attributes such as negation or uncertainty \citep{bao2019disentangledSyntaxSemantics,mercatali2021discreteTextVae,vasilakes2022negationUncertainty,zhang2024graphInducedVae}. The problem we study in this work is not directly addressed by this prior work. The key distinction is that these methods learn a latent variable model where the target distribution is not represented by the model at initialization. In our setting the base model already realizes $p_\theta(y|x)$, and it would be inefficient, if not impossible, to retrain the model from scratch to expose the strategy.

Mixture-of-experts language models also learn routing over specialized generators \citep{shazeer2017moe,jiang1999hme}, but such models are optimized for capacity or efficiency, not for correspondence to an interpretable strategy variable, and training is typically carried out  from scratch. Mechanistic interpretability methods, including sparse autoencoders \citep{galichin2025reasoningSaes} and concept-bottleneck approaches \citep{tan2024conceptBottlenecks}, extract or steer internal representations post-hoc; our goal is instead a generative factorization that supports sampling and intervention.

While there is also a rich theory for latent variable methods, classical identifiability theory for mixture models \citep{teicher1963mixtures,yakowitz1968mixtures,allman2009latentStructure} and nonlinear ICA \citep{hyvarinen2019nonlinearIca,khemakhem2020ivae} achieve uniqueness by restricting the component family or adding auxiliary observations. When using an expressive pretrained generator, classical identifiability  conditions generally fail, and the central question becomes how to select a useful factorization among the many that preserve the marginal distribution. An extended discussion of related work with fuller citations is given in the supplementary material.

\subsection{Implications and Future Directions}

The ideas and methods presented here open up several directions for future work; here we briefly mention potential connections to interpretability, directed exploration, and safety.
First, because a strategy-aligned factorization exposes the strategies implicit in a pretrained model's response distribution and makes them comparable across related inputs, it can provide a higher-level and more interpretable description of model behavior than token-level analysis. Moreover, intervening on $z$ at generation time can direct the model toward a specific strategy; shifting the router distribution $r_\phi(z\mid x)$ adjusts the relative weight of different strategies without retraining. This has the potential to provide a structured prior for exploring the strategy space, which could improve coverage in settings where the base model under-samples certain valid approaches. Finally, because strategy-level control offers a more interpretable abstraction for constraining or auditing generation compared with token-level steering, the methods introduced in this paper may support applications where certain reasoning paths need to be permitted or excluded, presenting possible benefits for safety.

The scope of our conclusions is limited in two respects.
The benchmarks we introduce cover synthetic algorithmic tasks with ground-truth strategies; how well the methodology extends to open-ended reasoning tasks remains an open question.
Moreover, the theoretical analysis is preliminary, and the connection between the oracle selection criterion and actual training dynamics is not yet formally characterized.

\section{Extended Related Work}
\label{app:related-work}

\paragraph{Conditional latent-variable models and language VAEs.}
Latent-variable generative models provide the formal background for our router-generator factorization.
The VAE and conditional VAE literature learns a latent-variable model from observed data, using an encoder or recognition model and an ELBO to train a generator with latent variation \citep{kingma2013vae,sohn2015cvae}.
Sequence and dialogue VAEs extend this idea to autoregressive generation, where the latent is intended to capture sentence-level, discourse-level, or response-level variation beyond the input context \citep{bowman2016sentenceVae,serban2016vhred,zhao2017dialogCvae}.
Pretrained language VAEs and latent-thought models scale related ideas with pretrained Transformer components, continuous sentence spaces, or inference-time latent optimization \citep{li2020optimus,park2021transformerVae,kong2025latentThoughts}.
Another line of language-VAE work explicitly structures the latent space around controllable or disentangled linguistic factors.
Prior work separates syntax and semantics in sentence VAEs \citep{bao2019disentangledSyntaxSemantics}, extends syntactic-semantic separation to Transformer-based VAEs \citep{zhang2024graphInducedVae}, uses discrete latent variables to capture natural-language generative factors \citep{mercatali2021discreteTextVae}, and disentangles semantic attributes such as negation and uncertainty from content \citep{vasilakes2022negationUncertainty}.
Latent-space control methods further model control distributions directly in the VAE latent space \citep{gu2023latentDensityControl}.
Our setting differs in the learning target: rather than fitting a latent-variable generator to data, we start from a fixed or pretrained response distribution $p_\theta(y|x)$ and learn a factorization $\int r_\phi(z|x)g_\phi(y|x,z)\,dz$ whose latent variable exposes strategy-level variation already implicit in that distribution.

\paragraph{Posterior collapse, latent usage, and disentanglement objectives.}
Autoregressive decoders can ignore latent variables, giving rise to posterior-collapse and KL-vanishing pathologies in sequence VAEs \citep{bowman2016sentenceVae,he2019laggingInference,fu2019cyclicalAnnealing}.
Several objective analyses separate the standard ELBO regularizer into latent information and aggregate-prior matching terms, motivating variants that preserve latent usage while keeping generation well regularized \citep{hoffman2016elboSurgery,zhao2019infovae}.
Disentanglement-oriented objectives such as $\beta$-VAE, capacity-controlled $\beta$-VAE, FactorVAE, and InfoGAN also show that useful latent structure usually requires explicit inductive bias or information pressure, not merely the presence of a latent code \citep{higgins2017betavae,burgess2018betavae,kim2018factorvae,chen2016infogan}.
These works motivate our use of objective terms that pressure $z$ to carry information.
However, our desideratum is more specific than latent activity or visual-factor disentanglement: $z$ should be informative about $y$ beyond $x$, aligned with reference solution strategies, and semantically consistent across related inputs.
The collapse issue is also sharper in our adaptation setting because the initialized generator already realizes the observable conditional behavior without consulting $z$.

\paragraph{Identifiability, conditional mixtures, and routing.}
Classical finite-mixture theory studies when a marginal distribution uniquely determines its mixture components, often through restrictive component families, linear-independence conditions, or multi-view structure \citep{teicher1963mixtures,yakowitz1968mixtures,allman2009latentStructure}.
Conditional mixture and mixture-of-experts models introduce input-dependent routing or gating, but their classical guarantees usually concern approximation, prediction, or consistency under structured expert families rather than recovery of a semantically named latent variable \citep{jiang1999hme}.
Identifiable latent-variable and nonlinear-ICA results restore stronger recovery guarantees by adding auxiliary variables or conditional-prior structure that breaks otherwise unavoidable symmetries \citep{hyvarinen2019nonlinearIca,khemakhem2020ivae}.
These literatures clarify why latent-variable factorization is not identified by marginal fit alone.
In our setting, the router and generator are expressive neural language models, so many factorizations can preserve $p_\theta(y|x)$.
The problem is therefore not only estimation of the observable conditional law, but selection of a useful factorization whose latent variable captures reusable strategy structure.

\paragraph{Interpretability and reasoning features.}
A separate line of work exposes internal model structure through supervised concept bottlenecks or post-hoc feature decompositions.
Concept-bottleneck methods for pretrained language models insert human-defined concept variables and train the model to predict through them, providing an explicit supervised interpretability interface \citep{tan2024conceptBottlenecks}.
Sparse-autoencoder analyses of reasoning models instead discover internal features associated with reasoning motifs and can steer model behavior by intervening on feature directions \citep{galichin2025reasoningSaes}.
These approaches are closely related to our interpretability motivation because they aim to make latent reasoning structure visible and manipulable.
The distinction is that we learn a generative router-generator factorization of $p_\theta(y|x)$, so the learned latent variable participates directly in generation rather than serving only as a supervised concept layer or a post-hoc activation feature.

\paragraph{Exploration and RLVR motivation.}
Recent analyses of reinforcement learning with verifiable rewards suggest that on-policy post-training often sharpens solutions already accessible under the base model rather than broadly expanding the set of reachable reasoning strategies \citep{yue2025rlvrBeyondBase,wu2025invisibleLeash}.
Representation-based exploration methods respond to this limitation by adding novelty or diversity pressure in model representation space \citep{tuyls2025representationExploration}.
This literature motivates strategy-level intervention as a possible future use of latent-variable factorization: a reusable strategy-aligned $z$ could provide a higher-level handle for directed exploration than token-space sampling alone.
We do not evaluate RLVR or downstream exploration in this paper; the connection is a motivation for why exposing abstract strategies in a pretrained model may be useful beyond the controlled benchmark.

\section{Benchmark and Data Generation}
\label{app:benchmark-data-generation}

The controlled multi-strategy algorithmic benchmark provides problem instances, solution traces, and reference strategy labels in a setting where strategy alignment can be measured directly.
Each task family is designed so that the same problem can be solved by multiple valid procedures.
The procedures differ in the intermediate solution trace while preserving the same final answer.
This gives an evaluation reference for asking whether a learned latent variable corresponds to strategy-level variation in generated reasoning traces.

\subsection{Task families}
\label{app:benchmark-task-families}

\Cref{tab:benchmark-task-families} summarizes the benchmark tasks used in the paper.
Within each task family, strategies are sampled uniformly from the listed strategy set.
The table lists the sampled strategies, not every procedure that could in principle solve the task.
The task families probe complementary forms of strategy variation: reduction order for summation, state-transition algorithms for sorting, move-order policies for grid paths, algebraic manipulation order for linear equations, conversion procedures for base conversion, and arithmetic decomposition for multidigit addition.
In each case, the strategies alter the intermediate trace while preserving the same problem-answer relation.

\begin{table}[h]
  \centering
  \small
  \caption{\textbf{Controlled multi-strategy algorithmic benchmark.}
  Each task family admits multiple valid solution procedures for the same problem instance.
  The benchmark uses the listed sampled strategies as reference labels for evaluating strategy-aligned latent variables.}
  \label{tab:benchmark-task-families}
  \setlength{\tabcolsep}{5pt}
  \renewcommand{\arraystretch}{1.08}
  \begin{tabular}{@{}p{0.22\linewidth}p{0.34\linewidth}p{0.36\linewidth}@{}}
    \toprule
    {\raggedright Task family\par}
      & {\raggedright Problem domain\par}
      & {\raggedright Sampled strategies\par} \tabularnewline
    \midrule
    {\raggedright List summation\par}
      & {\raggedright Four integers from \(0\) to \(9\)\par}
      & {\raggedright left-to-right; right-to-left; pairwise\par} \tabularnewline
    \addlinespace[0.25em]
    {\raggedright Sorting algorithms\par}
      & {\raggedright Five integers from \(0\) to \(9\)\par}
      & {\raggedright bubble sort; insertion sort; merge sort\par} \tabularnewline
    \addlinespace[0.25em]
    {\raggedright Grid pathfinding\par}
      & {\raggedright Monotone shortest paths on a \(6\times 6\) grid\par}
      & {\raggedright right-first; down-first; alternating\par} \tabularnewline
    \addlinespace[0.25em]
    {\raggedright Linear equation solving\par}
      & {\raggedright Integer equations \(ax+b=c\), with \(a=2,\ldots,9\) and integer-safe solutions and offsets in \([-9,9]\)\par}
      & {\raggedright subtract-then-divide; divide-then-subtract; inverse-ops\par} \tabularnewline
    \addlinespace[0.25em]
    {\raggedright Base conversion\par}
      & {\raggedright Integer \(n=1,\ldots,255\), base in \(2,4,8,16\)\par}
      & {\raggedright repeated division; via binary; decomposition\par} \tabularnewline
    \addlinespace[0.25em]
    {\raggedright Multidigit addition\par}
      & {\raggedright Two three-digit nonnegative integers\par}
      & {\raggedright right-to-left carry; left-to-right partials; rounding decomposition\par} \tabularnewline
    \bottomrule
  \end{tabular}
\end{table}

\subsection{Data generation}
\label{app:benchmark-data-generation-procedure}

The benchmark data distribution samples a task family, a problem instance for that task, and a reference strategy for solving that instance:
\[
  T \sim \mathrm{Unif}(\text{task families}),
  \qquad
  X\sim \mathcal D_T,
  \qquad
  S\sim \mathrm{Unif}(\mathcal S_T),
  \qquad
  Y=\mathrm{Trace}_T(X,S).
\]
Here \(\mathcal D_T\) is the task-specific input distribution and \(\mathcal S_T\) is the listed strategy set for task \(T\).
The rendering function \(\mathrm{Trace}_T\) produces a solution trace that follows strategy \(S\) and reaches the task's correct final answer.
The resulting example consists of a problem instance \(X\), a solution trace \(Y\), and a reference strategy label \(S\) that is used for evaluation but not training.

\subsection{Reference strategy identification}
\label{app:benchmark-strategy-identification}

The benchmark is designed to enable parsing solution trajectories generated by a model and identifying the reference strategy.
Given a problem instance and a generated solution trace, the task parser checks whether the trace is a valid solution and identifies which benchmark strategy it expresses.
This makes it possible to evaluate whether generated outputs preserve valid task behavior and whether a learned latent variable is aligned with the benchmark's reference strategy structure.
The reference labels are used as an evaluation target for the controlled benchmark.

\subsection{Representative examples}
\label{app:benchmark-representative-examples}

To illustrate the benchmark, the examples below show a sample problem from each task with two sample solutions demonstrating two different strategies.
Some traces are abbreviated with ellipses where the omitted steps continue the same procedure.

\paragraph{List summation.}
For the problem instance ``\texttt{3, 7, 2, 8}'', the benchmark can render traces such as:
\begin{center}
  \small
  \begin{tabular}{@{}p{0.24\linewidth}p{0.68\linewidth}@{}}
    \toprule
    Strategy & Trace \\
    \midrule
    left-to-right
      & \texttt{3+7=10 ; 10+2=12 ; 12+8=20} \\
    pairwise
      & \texttt{3+7=10 ; 2+8=10 ; 10+10=20} \\
    \bottomrule
  \end{tabular}
\end{center}

\paragraph{Sorting algorithms.}
For the problem instance ``\texttt{[5,2,8,1,4]}'', the benchmark can render traces such as:
\begin{center}
  \small
  \begin{tabular}{@{}p{0.24\linewidth}p{0.68\linewidth}@{}}
    \toprule
    Strategy & Trace \\
    \midrule
    bubble sort
      & \texttt{[2,5,8,1,4] ; [2,5,1,8,4] ; ... ; [1,2,4,5,8]} \\
    merge sort
      & \texttt{[2,5,8,1,4] ; [2,5,1,8,4] ; [1,2,5,8,4] ; [1,2,4,5,8]} \\
    \bottomrule
  \end{tabular}
\end{center}

\paragraph{Grid pathfinding.}
For the problem instance ``\texttt{grid=6 ; start=(0,0) ; goal=(2,3)}'', the benchmark can render traces such as:
\begin{center}
  \small
  \begin{tabular}{@{}p{0.24\linewidth}p{0.68\linewidth}@{}}
    \toprule
    Strategy & Trace \\
    \midrule
    right-first
      & \texttt{R:(0,1) ; R:(0,2) ; R:(0,3) ; D:(1,3) ; D:(2,3)} \\
    alternating
      & \texttt{R:(0,1) ; D:(1,1) ; R:(1,2) ; D:(2,2) ; R:(2,3)} \\
    \bottomrule
  \end{tabular}
\end{center}

\paragraph{Linear equation solving.}
For the problem instance ``\texttt{3x+6=21}'', the benchmark can render traces such as:
\begin{center}
  \small
  \begin{tabular}{@{}p{0.24\linewidth}p{0.68\linewidth}@{}}
    \toprule
    Strategy & Trace \\
    \midrule
    subtract-then-divide
      & \texttt{3x+6=21 ; 3x=15 ; x=5} \\
    divide-then-subtract
      & \texttt{3x+6=21 ; x+2=7 ; x=7-2 ; x=5} \\
    \bottomrule
  \end{tabular}
\end{center}

\paragraph{Base conversion.}
For the problem instance ``\texttt{n=45, base=4}'', the benchmark can render traces such as:
\begin{center}
  \small
  \begin{tabular}{@{}p{0.24\linewidth}p{0.68\linewidth}@{}}
    \toprule
    Strategy & Trace \\
    \midrule
    repeated division
      & \texttt{DIV:45/4=11,r:1 ; DIV:11/4=2,r:3 ; DIV:2/4=0,r:2 ; OUT:231} \\
    decomposition
      & \texttt{TERM:2*4\textasciicircum{}2=32 ; TERM:3*4\textasciicircum{}1=12 ; TERM:1*4\textasciicircum{}0=1 ; OUT:231} \\
    \bottomrule
  \end{tabular}
\end{center}

\paragraph{Multidigit addition.}
For the problem instance ``\texttt{348+274}'', the benchmark can render traces such as:
\begin{center}
  \small
  \begin{tabular}{@{}p{0.24\linewidth}p{0.68\linewidth}@{}}
    \toprule
    Strategy & Trace \\
    \midrule
    right-to-left carry
      & \texttt{d2:8+4+0=12->w2,c1 ; d1:4+7+1=12->w2,c1 ; d0:3+2+1=6->w6,c0 ; sum=622} \\
    left-to-right partials
      & \texttt{p100:300+200=500 ; p10:40+70=110 ; p1:8+4=12 ; sum=500+110+12=622} \\
    \bottomrule
  \end{tabular}
\end{center}

\section{Details of Training Methodology}\label{app:training-methodology-details}

\Cref{sec:methodology} describes the objective-level motivation for model-directed reconstruction.
This appendix gives the corresponding training procedure and the implementation details needed to reproduce the factorization objective.
Throughout this section, \(p_\theta\) is a frozen base model, \(\phi\) denotes the adapted router-generator parameters, and \(\xi\) denotes the training-time posterior parameters.
The training distribution is generated by the frozen base model itself: inputs are drawn from \(\calD_X\) and responses are sampled from \(p_\theta(\cdot|x)\).

\begin{algorithm}[t]
  \caption{Model-directed training of a router-generator factorization}
  \label{alg:model-directed-training}
  \DontPrintSemicolon
  \LinesNumbered
  \KwIn{Input distribution \(\calD_X\); frozen base model \(p_\theta\); objective parameters \(\beta,\alpha,\gamma\).}
  \KwOut{Router \(r_\phi(z|x)\) and strategy-conditioned generator \(g_\phi(y|x,z)\).}
  \vspace{0.5em}

  Initialize \(\phi\) from \(\theta\), adding router heads and a latent embedding projection\;
  Initialize \(\xi\) from \(\theta\), adding posterior heads\;
  Freeze \(p_\theta\) as both the response distribution to factorize and the reconstruction reference\;
  Estimate \(c_\theta = \bbE_{x\sim\calD_X,\;y\sim p_\theta(\cdot|x)}[b_\theta(x,y)]\) using frozen-base samples\;
  \For{each optimization step}{
    Sample a minibatch \(x_i\sim\calD_X\), for \(i=1,\ldots,B\)\;
    Sample responses \(y_i\sim p_\theta(\cdot|x_i)\)\;
    Compute frozen-base surprisals \(b_{\theta,t}(x_i,y_i)\) and weights \(a_{\theta,t}(x_i,y_i)\)\;
    Sample or reparameterize \(z_i\sim q_\xi(\cdot|x_i,y_i)\)\;
    Form \(\widehat{\calJ}_\theta=B^{-1}\sum_{i=1}^B\calJ_\theta(x_i,y_i;\phi,\xi)\)\;
    Update \(\phi\) and \(\xi\) by a stochastic gradient step on \(\widehat{\calJ}_\theta\), leaving \(\theta\) fixed\;
  }
\end{algorithm}

\subsection{Architecture Parameterization}
\label{app:training-methodology-details:architecture}

The router and generator are implemented as two roles of one adapted autoregressive Transformer.
The router reads the input prefix and produces a Gaussian distribution over continuous latent values.
Let \(h_{\phi}^{(L)}(x)\) be the final-layer representation produced by the adapted Transformer after reading \(x\).
The router distribution is
\[
  r_\phi(z|x)
  =
  \mathcal N\!\left(
    W_\mu h_{\phi}^{(L)}(x),
    \mathrm{diag}\!\left(\exp(W_\sigma h_{\phi}^{(L)}(x))\right)
  \right).
\]
During autoregressive generation, a latent sample \(z\) is projected to an embedding \(E_\phi(z)\) and inserted as a pseudo-token at the embedding layer.
The generator then predicts each response token while attending to \(x\), the latent pseudo-token, and the previous response tokens.

The training-time posterior uses the same construction, but reads the full input-response sequence.
Let \(h_{\xi}^{(L)}(x,y)\) be the final-layer representation after reading \((x,y)\).
The posterior distribution is
\[
  q_\xi(z|x,y)
  =
  \mathcal N\!\left(
    U_\mu h_{\xi}^{(L)}(x,y),
    \mathrm{diag}\!\left(\exp(U_\sigma h_{\xi}^{(L)}(x,y))\right)
  \right).
\]
The posterior is used only during training to provide variational signal.
At generation time, latents are sampled from the router.

Both adapted models are initialized from the frozen base model parameters \(\theta\).
In the experiments, the trainable parameters consist of LoRA-style low-rank updates to the Transformer backbone together with the Gaussian heads and latent embedding projection.
This keeps the factorization close to the base model while allowing the router and generator to assign a nontrivial role to \(z\).

\subsection{Objective and Normalization Details}
\label{app:training-methodology-details:objective}

For a sampled response \(y=(y_1,\ldots,y_{T_y})\), define the frozen-base token surprisal and response-average surprisal as
\[
  b_{\theta,t}(x,y)
  =
  -\log p_\theta(y_t|x,y_{<t}),
  \qquad
  b_\theta(x,y)
  =
  \frac{1}{T_y}
  \sum_{t=1}^{T_y} b_{\theta,t}(x,y).
\]
The global reference scale is
\[
  c_\theta
  =
  \bbE_{x\sim\calD_X,\;y\sim p_\theta(\cdot|x)}
  \left[
    b_\theta(x,y)
  \right].
\]
The \(\theta\) subscript marks dependence on the frozen base model.

Token-level reconstruction weights are derived from the base model surprisals.
For \(\gamma\ge0\), define
\[
  \widetilde w_{\theta,t}^{(\gamma)}(x,y)
  =
  \frac{
    b_{\theta,t}(x,y)^\gamma
  }{
    \sum_{s=1}^{T_y} b_{\theta,s}(x,y)^\gamma
  },
  \qquad
  \kappa_\theta^{(\gamma)}(x,y)
  =
  \frac{
    \sum_{s=1}^{T_y}
    \widetilde w_{\theta,s}^{(\gamma)}(x,y)
    b_{\theta,s}(x,y)
  }{
    b_\theta(x,y)
  },
\]
and
\[
  w_{\theta,t}^{(\gamma)}(x,y)
  =
  \frac{
    \widetilde w_{\theta,t}^{(\gamma)}(x,y)
  }{
    \kappa_\theta^{(\gamma)}(x,y)
  },
  \qquad
  a_{\theta,t}(x,y)
  =
  \alpha\,\frac{1}{T_y}
  +
  (1-\alpha)\,
  w_{\theta,t}^{(\gamma)}(x,y).
\]
The parameter \(\alpha\in[0,1]\) mixes uniform response-token pressure with token-level \(p_\theta\)-surprisal-directed pressure.
The parameter \(\gamma\) controls how strongly the token weights concentrate on high-surprisal positions.
The base model reference scores, token weights, and normalization constants defined above are treated as constants when updating \(\phi\) and \(\xi\).

The resulting reconstruction term and training objective are
\[
  \calR_\theta(x,y;\phi,\xi)
  =
  \frac{1}{c_\theta}
  \expectunder{z\sim q_\xi(\cdot|x,y)}{
    \sum_{t=1}^{T_y}
    a_{\theta,t}(x,y)
    \left[
      -\log g_\phi(y_t|x,z,y_{<t})
    \right]
  },
\]
and
\[
  \calJ_\theta(x,y;\phi,\xi)
  =
  \calR_\theta(x,y;\phi,\xi)
  +
  \beta\,
  \mathrm{KL}\!\left(
    q_\xi(\cdot|x,y)
    \,\Vert\,
    r_\phi(\cdot|x)
  \right).
\]
The normalization by \(\kappa_\theta^{(\gamma)}\) preserves the frozen-base response scale under token weighting:
\[
  \sum_{t=1}^{T_y}
  w_{\theta,t}^{(\gamma)}(x,y)\,
  b_{\theta,t}(x,y)
  =
  b_\theta(x,y).
\]
Since the uniform component also satisfies
\[
  \sum_{t=1}^{T_y}
  \frac{1}{T_y}
  b_{\theta,t}(x,y)
  =
  b_\theta(x,y),
\]
the mixed weights satisfy
\[
  \sum_{t=1}^{T_y}
  a_{\theta,t}(x,y)\,
  b_{\theta,t}(x,y)
  =
  b_\theta(x,y).
\]
Consequently, if the generator matches the frozen base model token distribution, so that \(g_\phi(y_t|x,z,y_{<t})=p_\theta(y_t|x,y_{<t})\), then \(\calR_\theta(x,y;\phi,\xi)=b_\theta(x,y)/c_\theta\).
Taking expectation over \(x\sim\calD_X\) and \(y\sim p_\theta(\cdot|x)\) gives \(\bbE[\calR_\theta]=1\).
Thus \(\calR_\theta\) can be read as a base-normalized fraction of the frozen-base response loss that remains after conditioning on the latent.

\subsection{Inference And Evaluation Interface}
\label{app:training-methodology-details:inference}

The posterior \(q_\xi(z|x,y)\) is a training-time object.
It provides latent samples for reconstruction and a variational bridge between the generator and router, but it is not needed to sample from the learned factorization.
At inference time, the model first samples \(z\sim r_\phi(\cdot|x)\) and then generates \(y\) autoregressively from \(g_\phi(\cdot|x,z)\).

Strategy annotations are not used in the optimization loop.
They enter only after training, when we evaluate whether the learned latent is informative and strategy-aligned, separates strategies within each input, and assigns semantically consistent latent regions across inputs

\section{Experimental Details}\label{app:experimental-details}

\subsection{Evaluation protocol and metrics}
\label{app:experimental-details:metrics}

We evaluate trained router-generator factorizations using the parseable structure of the benchmark described in~\Cref{app:benchmark-data-generation}.
Generated solution traces can be parsed and assigned to the benchmark's reference strategy structure, enabling direct evaluation of whether the learned latent variable preserves task behavior and has stable strategy meaning across related inputs.
Strategy labels are used only for evaluation.

\paragraph{Distributional Fidelity.}
Distributional Fidelity measures whether samples from the factorized model preserve the base model response behavior at the level of valid benchmark solutions.
Operationally, we sample \(Z\sim r_\phi(\cdot|X)\), generate \(Y\sim g_\phi(\cdot|X,Z)\), parse the generated trace, and report the fraction of samples that remain strategy-compatible task solutions.

\paragraph{Strategy Alignment.}
The main-paper Strategy Alignment metric is Analogical Consistency, as defined in~\Cref{sec:experiments:strategy-aligned-latents}.
We sample a source input \(X\), draw \(Z\sim r_\phi(\cdot|X)\), generate \(Y\sim g_\phi(\cdot|X,Z)\), reuse the same \(Z\) on a related target input \(X'\), and generate \(Y'\sim g_\phi(\cdot|X',Z)\).
The score is
\[
  \operatorname{AnalogicalConsistency}
  =
  \Pr\!\left[
    \operatorname{strat}(Y)
    =
    \operatorname{strat}(Y')
  \right],
\]
where \(X'\) is sampled from the same task family as \(X\).
This tests whether the latent value carries the same strategy meaning across different inputs.

\paragraph{Supporting diagnostics.}
Router Strategy Decodability is a linear-probe diagnostic measuring whether reference strategy information is linearly recoverable from router-sampled latents \(Z\sim r_\phi(\cdot|X)\).
It is useful for diagnosing latent information, but it is weaker than Analogical Consistency because it does not test whether the same latent value has reusable cross-input meaning.
For mechanism analysis, we also report base-relative reconstruction \(\calR_\theta\), token-weighted reconstruction, and the posterior-router prior KL term from the training objective.

\subsection{Objective variants and inverse-surprisal ablation}
\label{app:experimental-details:objectives}

\Cref{app:training-methodology-details:objective} gives the full model-directed objective.
The experiments compare variants of
\[
  \calJ(x,y;\phi,\xi)
  =
  \calR(x,y;\phi,\xi)
  +
  \beta\,
  \mathrm{KL}\!\left(
    q_\xi(\cdot|x,y)
    \,\Vert\,
    r_\phi(\cdot|x)
  \right),
\]
where the variants differ in the reconstruction term \(\calR\).
The standard ELBO uses the unnormalized autoregressive reconstruction loss, while the proposed variants use the frozen base model \(p_\theta\) to normalize and direct reconstruction pressure.

For uniform base-relative reconstruction, the token weights are \(a_{\theta,t}=1/T_y\).
For surprisal-proportionate token weighting, \(a_{\theta,t}\) is the concentration-corrected token weight defined from the frozen-base surprisal \(b_{\theta,t}(x,y)\) in~\Cref{app:training-methodology-details:objective}.
The global+token variants use convex mixtures of the uniform base-relative term and the high-surprisal token-weighted term.
\Cref{tab:experimental-details-objective-variants} maps the paper-facing method labels to the objective components used in the figures.

\begin{table}[t]
  \centering
  \small
  \caption{\textbf{Objective variants used in the experimental figures.}
  All model-directed variants train on samples \(x\sim\calD_X,\;y\sim p_\theta(\cdot|x)\) and keep \(p_\theta\) frozen as the reference model.}
  \label{tab:experimental-details-objective-variants}
  \setlength{\tabcolsep}{5pt}
  \renewcommand{\arraystretch}{1.08}
  \begin{tabular}{@{}p{0.19\linewidth}p{0.34\linewidth}p{0.37\linewidth}@{}}
    \toprule
    Method label & Reconstruction term & Role in comparison \\
    \midrule
    ELBO baseline
      & Standard token reconstruction, without base-relative normalization
      & Tests whether ordinary variational training identifies strategy-aligned latents. \\
    Global scale
      & Base-relative reconstruction with uniform token weights
      & Tests the effect of putting reconstruction in frozen-base reference units. \\
    Token weighting
      & Base-relative reconstruction with high-surprisal token weights
      & Tests model-directed pressure toward response regions where \(p_\theta\) has residual surprisal. \\
    Global + token
      & Convex mixture of uniform base-relative and high-surprisal token-weighted reconstruction
      & Combines global normalization with token-level model-directed pressure. \\
    Token inverse
      & Base-relative reconstruction with inverse-surprisal token weights
      & Directional negative control emphasizing low-surprisal tokens. \\
    \bottomrule
  \end{tabular}
\end{table}

\paragraph{Inverse-surprisal ablation.}
The inverse-surprisal ablation tests whether the direction of token-level pressure matters.
Let \(\tilde b_{\theta,t}(x,y)=\max\{b_{\theta,t}(x,y),\epsilon\}\), and define inverse-surprisal weights
\[
  \widetilde v_{\theta,t}(x,y)
  =
  \frac{
    \tilde b_{\theta,t}(x,y)^{-1}
  }{
    \sum_{s=1}^{T_y}
    \tilde b_{\theta,s}(x,y)^{-1}
  }.
\]
As with the high-surprisal token-weighted term, we use a per-example scale correction so that the z-free base model reconstruction remains on the same base-relative scale:
\[
  v_{\theta,t}(x,y)
  =
  \frac{
    \widetilde v_{\theta,t}(x,y)
  }{
    \eta_\theta(x,y)
  },
  \qquad
  \eta_\theta(x,y)
  =
  \frac{
    \sum_{s=1}^{T_y}
    \widetilde v_{\theta,s}(x,y)
    b_{\theta,s}(x,y)
  }{
    b_\theta(x,y)
  }.
\]
Replacing \(a_{\theta,t}\) by \(v_{\theta,t}\) emphasizes tokens already well predicted by \(p_\theta\), while preserving the same z-free reconstruction scale in expectation.
The ablation therefore tests the mechanism-level hypothesis that alignment gains come from pressure on high-surprisal, strategy-relevant response regions rather than from nonuniform token weighting alone.

\subsection{Training settings and hyperparameter sweeps}
\label{app:experimental-details:sweeps}

The default experimental setting is multi-task factorization: each run trains one router-generator model jointly across all six benchmark task families.
This setting better matches the heterogeneous training regime of language models, where a single model must represent many task distributions.
Additional single-task runs are used for diagnostic sweeps and mechanism plots where per-task variation is useful.

We construct the benchmark base models from two initializations: pretrained Qwen2.5 checkpoints and random weights.
In both settings, we first fit the model to the benchmark task distribution; we then freeze and factorize the resulting base model \(p_\theta\).
The randomly initialized diagnostic grid contains 504 complete LoRA adaptation runs: six single-task settings plus the multi-task setting, twelve reconstruction-family settings including ablations, three KL weights \(\beta\in\{1,0.1,0.01\}\), and two schedules for \(\beta\) (constant and linear warmup over training).
The pretrained method-comparison runs factorize Qwen2.5 0.5B and 1.5B base models~\citep{qwen25technicalreport} in the multi-task setting with continuous latent dimension 64.
We use the Hugging Face checkpoints \texttt{Qwen/Qwen2.5-0.5B} and \texttt{Qwen/Qwen2.5-1.5B}, whose model cards list the Apache 2.0 license.

\subsection{Additional diagnostics}
\label{app:experimental-details:additional-diagnostics}

The main text focuses on the headline pretrained method comparison and compact mechanism diagnostics.
Here we include additional views that support the same claims: latent geometry over training, role of KL weights, metric associations, and phase-space training dynamics.

\begin{figure}[t]
  \centering
  \includegraphics[width=\textwidth]{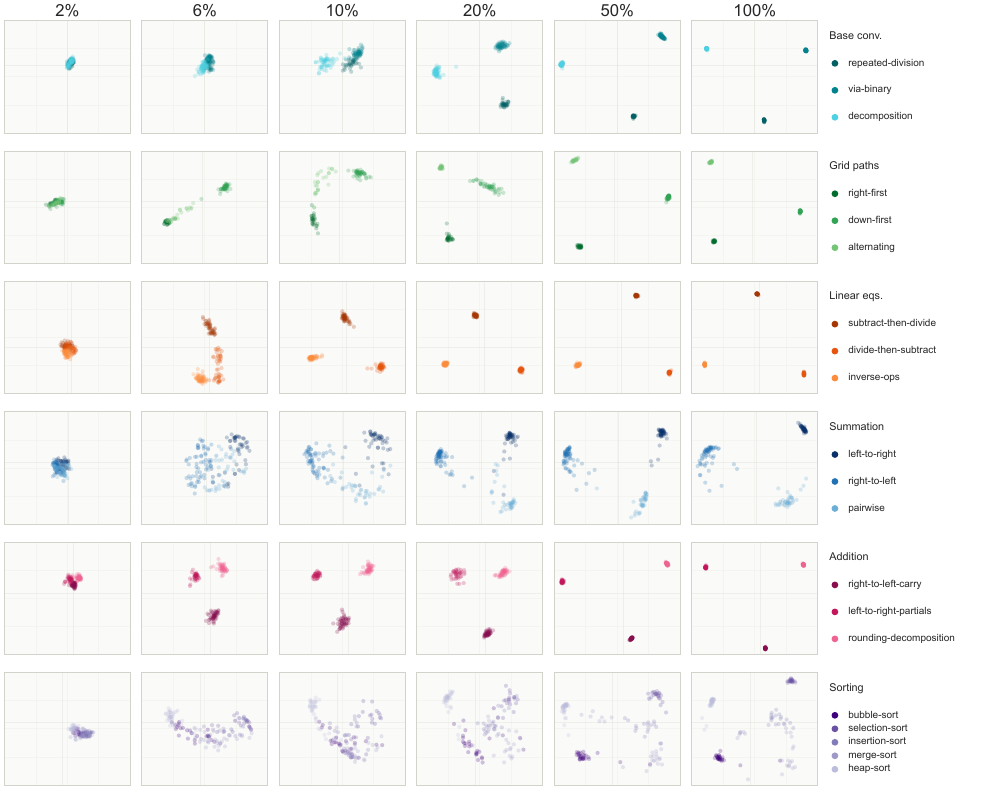}
  \caption{\textbf{Task-faceted posterior latent dynamics.}
  Each row shows one benchmark task for the selected multi-task token-weighting run, and columns show checkpoints over training.
  Points are posterior mean latents projected with task-local PCA and colored by reference strategy.
  The view complements \Cref{fig:disentanglement} by showing that strategy organization emerges within each task family.}
  \label{fig:appendix-posterior-by-task-static}
\end{figure}

\begin{figure}[t]
  \centering
  \includegraphics[width=\textwidth]{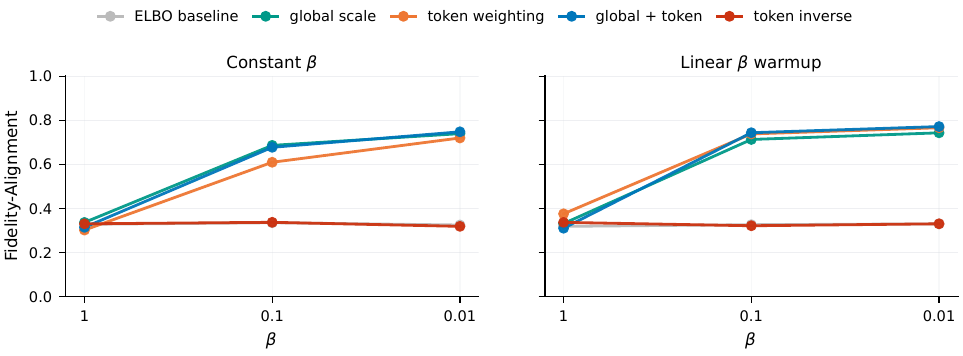}
  \caption{\textbf{Method success across \(\beta\) settings.}
  Plots show the product of Distributional Fidelity and Strategy Alignment across different KL weights \(\beta\) for the multi-task randomly initialized diagnostic grid.
  Standard ELBO-style baselines remain weak in fidelity-alignment across \(\beta\), while model-directed objectives succeed robustly once \(\beta\) is small enough to enable informative latents.}
  \label{fig:appendix-fidelity-alignment-beta-robustness}
\end{figure}

\begin{figure}[t]
  \centering
  \begin{minipage}[t]{0.49\textwidth}
    \centering
    \includegraphics[width=\linewidth]{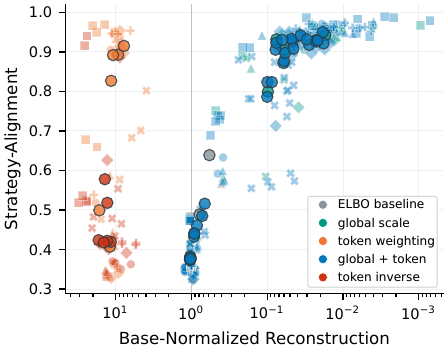}
  \end{minipage}\hfill
  \begin{minipage}[t]{0.49\textwidth}
    \centering
    \includegraphics[width=\linewidth]{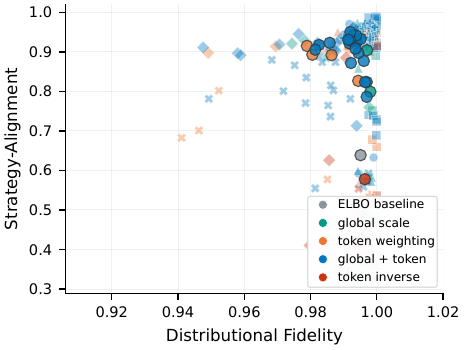}
  \end{minipage}
  \caption{\textbf{Endpoint metric associations in single-task diagnostic runs.}
  \textbf{Left:} base-relative reconstruction versus Strategy Alignment. There is a general positive association between reconstruction measured in base model units and strategy alignment; however, unlike the token-weighted reconstruction shown in the main text, the association is weaker for the pure token-weighted variant. This is likely caused by too little reconstruction pressure on non high-surprisal tokens, which causes some loss on these tokens. This motivates setting $\alpha > 0$ in the global+token variant to preserve some uniform reconstruction pressure.
  \textbf{Right:} Distributional Fidelity versus Strategy Alignment.
  Larger outlined points show method centroids and smaller translucent points show individual runs; these diagnostics illustrate that it is possible to have high distributional fidelity without strategy alignment. In general distributional fidelity is easy, especially because the factorization is initialized near the pretrained model's response distribution, but strategy alignment is harder and requires the right objective design.}
  \label{fig:appendix-base-reconstruction-strategy-alignment-scatter}
  \label{fig:appendix-fidelity-strategy-alignment-scatter}
\end{figure}

\begin{figure}[t]
  \centering
  \includegraphics[width=\textwidth]{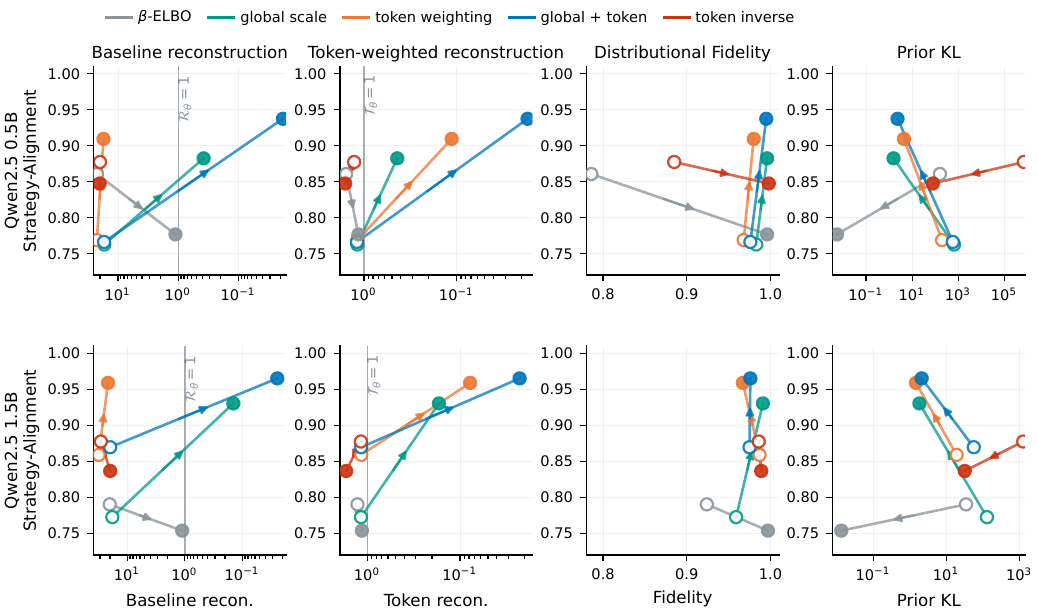}
  \caption{\textbf{Phase-space training dynamics of pretrained Qwen model factorization.}
  Phase-space dynamics diagrams of Qwen2.5 0.5B and 1.5B multi-task runs are shown across base-relative reconstruction, token-weighted reconstruction, Distributional Fidelity, and Prior KL, each plotted against Strategy Alignment.
  Trajectories show how each method moves through these diagnostic phase spaces over training, from initialization to the final checkpoint.}
  \label{fig:appendix-pretrained-phase-space-grid}
\end{figure}

\section{A Theory of Variational Approaches to the Strategy-Structure Recovery Problem}
\label{app:variational-complexity-theory}

This appendix gives the formal treatment behind the preview theorem in the main text.
The main result studies the population ELBO over independently parameterized expressive encoder, router, and generator classes.
% The hard deterministic attribute calculation is recovered as a specialization when the profiled router-generator optimum is the exact induced factorization.

\subsection{Setup and the Strategy Structure Recovery Problem}
\label{app:setup}

Let \(\calX\) and \(\calY\) be finite sets (e.g., space of token sequences).
Let \(\mu\) be a fixed input law on \(\calX\), and let \(p(y\mid x)\) be a fixed conditional law on \(\calY\).
All expectations, entropies, and mutual informations are computed under
\[
  X\sim\mu, \qquad Y\sim p(\cdot\mid X),
\]
unless stated otherwise.
% All logarithms are natural. % is this needed?
Because all sample spaces are finite, measurability is automatic; the only analytic issues are support, finite-value, and minimizer-attainment issues, stated explicitly below.

Fix a latent cardinality \(K\), and write \(\calZ=[K]\).
Let \(\calQ,\calR,\calG\) be classes of encoders \(q(z\mid x,y)\), routers \(r(z\mid x)\), and generators \(g(y\mid x,z)\), respectively.
For \(q\in\calQ\), define the joint law
\[
  Q_q(x,y,z)=\mu(x)p(y\mid x)q(z\mid x,y).
\]
Under \(Q_q\), define the encoder-induced router
\[
  q(z\mid x):=\sum_y p(y\mid x)q(z\mid x,y),
\]
and the encoder-induced branch conditional, whenever \(q(z\mid x)>0\), by
\[
  p_q(y\mid x,z) := \frac{p(y\mid x)q(z\mid x,y)}{q(z\mid x)}.
\]
The conditional \(p_q(y\mid x,z)\) may be defined arbitrarily on zero-mass branches; all KL and expectation terms below weight such branches by \(q(z\mid x)\).
Let \(I_q(Y;Z\mid X)\) denote conditional mutual information under \(Q_q\).

Let \(S^\star=\sigma(X,Y)\in[K]\) denote the reference strategy.
This hard-strategy assumption says that, once the completed trace is observed, the strategy is recognizable as an attribute of \((X,Y)\).
Let \(q^\star\) be the hard reference encoder
\[
  q^\star(z\mid x,y):=\Ind{z=\sigma(x,y)}.
\]
We assume \(q^\star\in\calQ\) when stating recovery theorems.

Because latent labels are arbitrary, recovery can only be defined up to relabeling.
The next definition separates relabelings that may vary with the input from a single label system shared across inputs.

\begin{definitionbox}[Weak and strong recovery]
An encoder \(q\in\calQ\) weakly recovers \(S^\star\) if, for \(\mu\)-almost every \(x\), there exists a permutation \(\pi_x\) of \([K]\) such that
\[
  Z=\pi_x(S^\star) \qquad Q_q\text{-almost surely conditional on }X=x.
\]
It strongly recovers \(S^\star\) if there exists one global permutation \(\pi\) of \([K]\) such that \(Z=\pi(S^\star)\) \(Q_q\)-almost surely.
Null labels are ignored; any bijection between positive-probability labels may be extended arbitrarily to a permutation of \([K]\).
\end{definitionbox}

We measure approximate recovery by the corresponding variation-of-information distances under \(Q_q\):
\[
  d_{\mathrm w}(q,S^\star) := H_q(Z\mid X,S^\star)+H_q(S^\star\mid X,Z),
\]
and
\[
  d_{\mathrm s}(q,S^\star) := H_q(Z\mid S^\star)+H_q(S^\star\mid Z).
\]
The weak distance conditions on \(X\), so it compares strategy semantics separately within each input.
The strong distance does not condition on \(X\), so it also penalizes input-dependent relabeling.
The following lemma verifies that these distances are exactly the right zero-error notions for the two definitions.

\begin{lemmabox}[Recovery distances]
\label{lem:recovery-distances}
For every \(q\in\calQ\):
\begin{enumerate}
\item \(d_{\mathrm w}(q,S^\star)=0\) if and only if \(q\) weakly recovers \(S^\star\).
\item \(d_{\mathrm s}(q,S^\star)=0\) if and only if \(q\) strongly recovers \(S^\star\).
\item \(d_{\mathrm w}(q,S^\star)\le d_{\mathrm s}(q,S^\star)\).
\end{enumerate}
\end{lemmabox}

\begin{proof}
For the weak statement, \(d_{\mathrm w}(q,S^\star)=0\) holds if and only if
\[
  H_q(Z\mid X,S^\star)=0, \qquad H_q(S^\star\mid X,Z)=0.
\]
Thus, for \(\mu\)-almost every \(x\), \(Z\) is a deterministic function of \(S^\star\) under \(Q_q(\cdot\mid X=x)\), and \(S^\star\) is a deterministic function of \(Z\) under the same conditional law.
The two variables therefore induce the same partition of positive-probability labels at that input, up to a bijection between the labels that occur.
Since both label spaces are subsets of \([K]\), the bijection extends to a permutation \(\pi_x\).
This is weak recovery.
The converse is immediate because \(Z=\pi_x(S^\star)\) makes the two conditional entropies vanish.

The strong statement is the same argument without conditioning on \(X\).
The condition \(d_{\mathrm s}(q,S^\star)=0\) is equivalent to \(Z\) and \(S^\star\) determining each other as global random variables under \(Q_q\), which is strong recovery up to null labels.

Finally, conditioning reduces entropy:
\[
  H_q(Z\mid X,S^\star)\le H_q(Z\mid S^\star), \qquad H_q(S^\star\mid X,Z)\le H_q(S^\star\mid Z).
\]
Adding these inequalities gives \(d_{\mathrm w}(q,S^\star)\le d_{\mathrm s}(q,S^\star)\).
\end{proof}

Thus any strong recovery guarantee also implies weak recovery, but not conversely.
This distinction becomes important when different inputs can reuse the same latent labels with different meanings.

\subsection{ELBO Decomposition with Router-Generator Complexity}
\label{app:information-complexity-reduction}

We study the population complexity-penalized beta-ELBO, with \(\beta\ge0\) and \(\lambda\ge0\),
\[
  \calL_{\beta,\lambda}(q,r,g) := \Expect_{Q_q}[-\log g(Y\mid X,Z)] + \beta\,\Expect_{X,Y}\mathrm{KL}(q(\cdot\mid X,Y)\|r(\cdot\mid X)) + \lambda C(r,g).
\]
The functional \(C(r,g)\in[0,\infty]\) is a theoretical proxy for architecture, parameterization, and optimization biases that favor simple router-generator factorizations.
We assume it is invariant under global relabeling of latent states.
All KL terms use the usual extended-real convention.

Define
\[
  \DecGap(q,g) := \Expect_{X,Z} \mathrm{KL}(p_q(\cdot\mid X,Z)\|g(\cdot\mid X,Z)),
\]
and
\[
  \PriorGap(q,r) := \Expect_X \mathrm{KL}(q(\cdot\mid X)\|r(\cdot\mid X)).
\]

The following lemma decomposes the variational objective into four terms: the irreducible conditional entropy \(H(Y\mid X)\), an information term involving \(I_q(Y;Z\mid X)\), the router and decoder KL gaps, and the complexity penalty \(C(r,g)\).
This form makes clear how the objective trades off information in the latent variable against the fit and complexity of the router-generator factorization.

\begin{lemmabox}[ELBO information-complexity decomposition]
\label{lem:information-complexity-reduction}
For every \((q,r,g)\),% for which the terms are well-defined,
\[
\begin{aligned}
  \calL_{\beta,\lambda}(q,r,g) &={} H(Y\mid X) + (\beta-1)I_q(Y;Z\mid X) \\ &\quad+ \DecGap(q,g) + \beta\PriorGap(q,r) + \lambda C(r,g).
\end{aligned}
\]
\end{lemmabox}

\begin{proof}
First decompose the reconstruction term.
Under \(Q_q\),
\[
  \Expect[-\log g(Y\mid X,Z)] = \Expect_{X,Z}\sum_y p_q(y\mid X,Z)(-\log g(y\mid X,Z)).
\]
For each \((x,z)\), cross-entropy equals entropy plus KL:
\[
  \sum_y p_q(y\mid x,z)(-\log g(y\mid x,z)) = H(p_q(\cdot\mid x,z)) + \mathrm{KL}(p_q(\cdot\mid x,z)\|g(\cdot\mid x,z)).
\]
Averaging gives
\[
  \Expect[-\log g(Y\mid X,Z)] = H_q(Y\mid X,Z)+\DecGap(q,g).
\]
Because the \((X,Y)\) marginal of \(Q_q\) is \(\mu(x)p(y\mid x)\),
\[
  H_q(Y\mid X)=H(Y\mid X).
\]
Thus
\[
  H_q(Y\mid X,Z)=H(Y\mid X)-I_q(Y;Z\mid X),
\]
and
\[
  \Expect[-\log g(Y\mid X,Z)] = H(Y\mid X)-I_q(Y;Z\mid X)+\DecGap(q,g).
\]

Next decompose the KL-to-prior term:
\[
  \Expect_{X,Y}\mathrm{KL}(q(\cdot\mid X,Y)\|r(\cdot\mid X)) = \Expect_{Q_q}\log\frac{q(Z\mid X,Y)}{r(Z\mid X)}.
\]
Insert \(q(Z\mid X)\):
\[
  \log\frac{q(Z\mid X,Y)}{r(Z\mid X)} = \log\frac{q(Z\mid X,Y)}{q(Z\mid X)} + \log\frac{q(Z\mid X)}{r(Z\mid X)}.
\]
The first expectation is \(I_q(Y;Z\mid X)\), and the second is \(\PriorGap(q,r)\).
Therefore
\[
  \Expect_{X,Y}\mathrm{KL}(q(\cdot\mid X,Y)\|r(\cdot\mid X)) = I_q(Y;Z\mid X)+\PriorGap(q,r).
\]
Combining the reconstruction decomposition, the KL decomposition, and the complexity term gives the identity.
\end{proof}

The coefficient of \(I_q(Y;Z\mid X)\) is \(\beta-1\).
Therefore the objective is information-seeking only when \(\beta<1\).
At \(\beta=1\), selection can still occur through the profiled cost below, but the ELBO no longer rewards conditional information after the variational gaps have been profiled out.
For \(\beta>1\), the objective penalizes information in \(Z\).

\subsection{Profiled Router-Generator Objective}
\label{app:profiled-objective}

For a fixed encoder \(q\), define the profiled router-generator cost
\[
  A_{\beta,\lambda}(q) := \inf_{r\in\calR,\;g\in\calG} \left[ \DecGap(q,g)+\beta\PriorGap(q,r)+\lambda C(r,g) \right] \in[0,\infty].
\]
Also define the profiled population loss
\[
  \overline{\calL}_{\beta,\lambda}(q) := \inf_{r\in\calR,\;g\in\calG} \calL_{\beta,\lambda}(q,r,g).
\]

Profiling asks how good an encoder can look after choosing the best available router and generator for it.
The next proposition rewrites that profiled loss as an encoder-only score.

\begin{propositionbox}[Profiled ELBO score]
\label{prop:profiled-elbo-score}
For every \(q\in\calQ\), %in the extended-real sense,
\[
  \overline{\calL}_{\beta,\lambda}(q) = H(Y\mid X) - (1-\beta)I_q(Y;Z\mid X) + A_{\beta,\lambda}(q).
\]
If \(\beta<1\), define
\[
  J^{\mathrm{prof}}_{\beta,\lambda}(q) := I_q(Y;Z\mid X) - \frac{A_{\beta,\lambda}(q)}{1-\beta},
\]
% with \(J^{\mathrm{prof}}_{\beta,\lambda}(q)=-\infty\) when \(A_{\beta,\lambda}(q)=\infty\).
Then
\[
  \overline{\calL}_{\beta,\lambda}(q) = H(Y\mid X) - (1-\beta)J^{\mathrm{prof}}_{\beta,\lambda}(q).
\]
\end{propositionbox}

\begin{proof}
Take the infimum over \(r\in\calR\) and \(g\in\calG\) in Lemma~\ref{lem:information-complexity-reduction}.
The terms \(H(Y\mid X)\) and \(I_q(Y;Z\mid X)\) do not depend on \(r\) or \(g\), and the remaining infimum is exactly \(A_{\beta,\lambda}(q)\).
The final identity is a rearrangement when \(\beta<1\).
\end{proof}

Thus the full expressive ELBO selects encoders by an information-minus-profiled approximation-complexity score.
The quantity \(A_{\beta,\lambda}(q)\), which involves the router-generator complexity \(C(r, g)\), measures whether the encoder-induced routing law and branch conditionals can be approximated with low profiled cost by a simple router-generator pair.

\subsection{Metric Recovery Margins}
\label{app:metric-recovery}

Fix a notion of recovery \(\rho\in\{\mathrm w,\mathrm s\}\) and \(\delta>0\).
Define the bad encoder set
\[
  \calB_\rho(\delta) := \{q\in\calQ:d_\rho(q,S^\star)\ge\delta\}.
\]
For \(\beta<1\) and \(A_{\beta,\lambda}(q^\star)<\infty\), define the profiled metric recovery margin
\[
  \Delta_{\beta,\lambda}^{\rho,\mathrm{prof}}(\delta) := J^{\mathrm{prof}}_{\beta,\lambda}(q^\star) - \sup_{q\in\calB_\rho(\delta)} J^{\mathrm{prof}}_{\beta,\lambda}(q),
\]
with the convention that the supremum of an empty set is \(-\infty\).
Equivalently,
\[
  \Delta_{\beta,\lambda}^{\rho,\mathrm{prof}}(\delta) = \inf_{q\in\calB_\rho(\delta)} \left[ I^\star-I(q) +
    \frac{A_{\beta,\lambda}(q)-A_{\beta,\lambda}(q^\star)}{1-\beta}
  \right],
\]
where \(I(q):=I_q(Y;Z\mid X)\) and \(I^\star:=I_{q^\star}(Y;Z\mid X)\).
% This display is only a shorthand when the bad set is nonempty; the previous definition is primary.

The following theorem states the population strategy-structure recovery criterion.
When the profiled margin at level \(\delta\) is positive, every population ELBO minimizer must recover the reference strategy up to recovery distance \(\delta\).

\begin{theorembox}[Population strategy-structure recovery for the expressive ELBO]
\label{thm:population-profiled-metric-recovery}
Assume \(\beta<1\), \(q^\star\in\calQ\), \(A_{\beta,\lambda}(q^\star)<\infty\), and \(\Delta_{\beta,\lambda}^{\rho,\mathrm{prof}}(\delta)>0\).
If a profiled minimizer exists and
\[
  \widehat q \in \argmin_{q\in\calQ}\overline{\calL}_{\beta,\lambda}(q),
\]
then
\[
  d_\rho(\widehat q,S^\star)<\delta.
\]
Consequently, if a full ELBO minimizer exists and
\[
  (\widehat q,\widehat r,\widehat g) \in \argmin_{q\in\calQ,\;r\in\calR,\;g\in\calG} \calL_{\beta,\lambda}(q,r,g),
\]
then \(d_\rho(\widehat q,S^\star)<\delta\).
\end{theorembox}

\begin{proof}
By Proposition~\ref{prop:profiled-elbo-score}, minimizing \(\overline{\calL}_{\beta,\lambda}\) is equivalent to maximizing \(J^{\mathrm{prof}}_{\beta,\lambda}\) when \(\beta<1\).
Hence any profiled population minimizer \(\widehat q\) satisfies
\[
  J^{\mathrm{prof}}_{\beta,\lambda}(\widehat q) \ge J^{\mathrm{prof}}_{\beta,\lambda}(q^\star).
\]
If \(d_\rho(\widehat q,S^\star)\ge\delta\), then \(\widehat q\in\calB_\rho(\delta)\), so
\[
  J^{\mathrm{prof}}_{\beta,\lambda}(\widehat q) \le \sup_{q\in\calB_\rho(\delta)} J^{\mathrm{prof}}_{\beta,\lambda}(q) = J^{\mathrm{prof}}_{\beta,\lambda}(q^\star) - \Delta_{\beta,\lambda}^{\rho,\mathrm{prof}}(\delta) < J^{\mathrm{prof}}_{\beta,\lambda}(q^\star),
\]
contradicting optimality.
Therefore \(d_\rho(\widehat q,S^\star)<\delta\).

For the final claim, let \((\widehat q,\widehat r,\widehat g)\) be a full ELBO minimizer and let
\[
  V:=\inf_{q\in\calQ,\,r\in\calR,\,g\in\calG}\calL_{\beta,\lambda}(q,r,g) = \calL_{\beta,\lambda}(\widehat q,\widehat r,\widehat g).
\]
For every \(q\), \(V\le\overline{\calL}_{\beta,\lambda}(q)\), so \(V\le\inf_q\overline{\calL}_{\beta,\lambda}(q)\).
Conversely,
\[
  \inf_q\overline{\calL}_{\beta,\lambda}(q) \le \overline{\calL}_{\beta,\lambda}(\widehat q) \le \calL_{\beta,\lambda}(\widehat q,\widehat r,\widehat g) =V.
\]
Thus \(\widehat q\) minimizes the profiled loss over \(\calQ\), and the first part applies.
\end{proof}

This theorem reduces population recovery to a score-separation condition.
It states that semantic recovery of the reference strategy is controlled by the profiled-score gap \(\Delta_{\beta,\lambda}^{\rho,\mathrm{prof}}(\delta)\):
if this gap is positive, then optimizing the complexity-penalized variational objective selects only encoders inside the desired \(\delta\)-recovery neighborhood.

\begin{remarkbox}[Sufficient conditions for minimizer existence]
  The above theorem is conditional on minimizers existing.
  One sufficient route, in these finite ambient spaces, is that \(\calQ,\calR,\calG\) are compact subsets of the relevant probability simplexes and \(C\) is lower semicontinuous, with support restrictions or closure conventions that make the KL terms lower semicontinuous and avoid undefined \(\infty-\infty\) expressions.
  The substantive content of this theorem is the identification of the profiled score that must separate the target encoder from misaligned encoders.
\end{remarkbox}

The margin definition is sometimes hard to check directly because it ranges over all bad encoders.
The following corollary gives a more interpretable sufficient condition in terms of an information advantage and a profiled-cost advantage. It says that if encoders that fail to recover the strategy don't have much more conditional information than the target, and they pay a large enough profiled cost penalty, then the selected encoder will recover the strategy.

\begin{corollarybox}[A sufficient information-complexity separation condition]
\label{cor:sufficient-margin-condition}
Assume the hypotheses of Theorem~\ref{thm:population-profiled-metric-recovery} except for the positivity of the margin.
Suppose there exist \(\gamma_I\in\mathbb R\) and \(\gamma_A>0\) such that, for every \(q\in\calB_\rho(\delta)\),
\[
  I_q(Y;Z\mid X)-I_{q^\star}(Y;Z\mid X) \le \gamma_I,
\]
and
\[
  A_{\beta,\lambda}(q)-A_{\beta,\lambda}(q^\star) \ge \gamma_A.
\]
If \(\gamma_A>(1-\beta)\gamma_I\), then
\[
  \Delta_{\beta,\lambda}^{\rho,\mathrm{prof}}(\delta) \ge \frac{\gamma_A}{1-\beta}-\gamma_I >0.
\]
\end{corollarybox}

\begin{proof}
For every bad \(q\),
\[
  J^{\mathrm{prof}}_{\beta,\lambda}(q^\star) - J^{\mathrm{prof}}_{\beta,\lambda}(q) = I_{q^\star}(Y;Z\mid X)-I_q(Y;Z\mid X) +
  \frac{A_{\beta,\lambda}(q)-A_{\beta,\lambda}(q^\star)}{1-\beta}
  \ge -\gamma_I+\frac{\gamma_A}{1-\beta}.
\]
Taking the infimum over \(\calB_\rho(\delta)\) proves the claim.
\end{proof}

This corollary makes the tradeoff explicit: a bad encoder may have more conditional information, but only up to \(\gamma_I\).
The target still wins if the bad encoder pays a large enough additional router-generator cost.

We now return to the distinction between weak and strong recovery.
Strong recovery requires more than recovering the correct partition separately for each input; it also requires label alignment \emph{across} inputs.
The next proposition shows that input-dependent relabelings can preserve weak recovery and conditional information exactly, while still preventing strong recovery unless the profiled cost favors a globally consistent label system.

% \aanote{Can this proposition be stated in a more structured and clear way? E.g., a list of statments.}
\begin{propositionbox}[Input-dependent relabelings and strong recovery]
\label{prop:input-dependent-relabeling}
For a family of permutations \(\pi=\{\pi_x:x\in\calX\}\), define the hard encoder
\[
  q^\pi(z\mid x,y):=\Ind{z=\pi_x(\sigma(x,y))}.
\]
Assume \(\beta<1\).
If \(q^\pi\in\calQ\), then \(d_{\mathrm w}(q^\pi,S^\star)=0\) and
\[
  I_{q^\pi}(Y;Z\mid X)=I_{q^\star}(Y;Z\mid X).
\]
If no single global permutation agrees with \(\pi_x\) on the positive-probability support of \(S^\star\mid X=x\) for \(\mu\)-almost every \(x\), then \(d_{\mathrm s}(q^\pi,S^\star)>0\).
In that case, \(q^\star\) beats this input-dependent relabeling in the profiled score if and only if
\[
  A_{\beta,\lambda}(q^\pi)>A_{\beta,\lambda}(q^\star).
\]
Consequently, strong recovery requires a positive profiled-cost gap against such non-globally aligned relabelings; if some such \(q^\pi\) has \(A_{\beta,\lambda}(q^\pi)=A_{\beta,\lambda}(q^\star)\), then the strong margin cannot be positive for every \(\delta\le d_{\mathrm s}(q^\pi,S^\star)\).
\end{propositionbox}

\begin{proof}
Conditional on \(X=x\), the map \(S^\star\mapsto\pi_x(S^\star)\) is a bijection on labels, so weak recovery holds and
\[
  I_{q^\pi}(Y;Z\mid X)=H_{q^\pi}(Z\mid X)=H(S^\star\mid X) =I_{q^\star}(Y;Z\mid X).
\]
If no global permutation agrees with the family \(\{\pi_x\}\) on the positive supports, then \(q^\pi\) does not strongly recover \(S^\star\); by Lemma~\ref{lem:recovery-distances}, \(d_{\mathrm s}(q^\pi,S^\star)>0\).
Since the conditional information terms are equal, the profiled-score comparison between \(q^\star\) and \(q^\pi\) is determined exactly by the profiled costs.
If the profiled costs tie, the two scores tie, so no positive strong margin can separate \(q^\star\) from \(q^\pi\) at any scale including \(q^\pi\) in the bad set.
\end{proof}

This shows why strong recovery requires more than predictive information.
The model class or complexity proxy must also prefer a globally aligned label system over equally informative local relabelings.

The recovery theorem above gives approximate recovery at a chosen radius with respect to the corresponding recovery distance.
When the candidate class has a positive distance gap around the target equivalence class, that approximate statement upgrades to exact recovery.

\begin{corollarybox}[Exact recovery from metric recovery]
\label{cor:exact-recovery}
Let \(\sim_\rho\) denote weak equivalence when \(\rho=\mathrm w\) and strong equivalence when \(\rho=\mathrm s\).
Suppose
\[
  m_\rho:=\inf_{q\not\sim_\rho q^\star}d_\rho(q,S^\star)>0.
\]
If Theorem~\ref{thm:population-profiled-metric-recovery} holds for some \(\delta\le m_\rho\), then every population ELBO minimizer exactly recovers \(S^\star\) in the corresponding weak or strong sense.
\end{corollarybox}

\begin{proof}
The theorem gives \(d_\rho(\widehat q,S^\star)<\delta\le m_\rho\).
By definition of \(m_\rho\), every non-recovering encoder has distance at least \(m_\rho\).
Therefore \(\widehat q\sim_\rho q^\star\).
\end{proof}

The separation assumption \(m_\rho>0\) is natural for finite or otherwise discrete encoder classes.
% It is typically false for rich convex stochastic classes: mixtures \((1-\varepsilon)q^\star+\varepsilon q_{\mathrm{bad}}\) can be non-recovering while their recovery distance tends to zero with \(\varepsilon\).

\begin{examplebox}[Positive margin from semantic branch sharing]
Suppose the target strategy yields branch conditionals that can be represented by shared low-complexity generator templates, while every encoder at recovery scale \(\delta\) either has at most \(\gamma_I\) more conditional information or requires at least \(\gamma_A\) additional profiled router-generator cost.
If \(\gamma_A>(1-\beta)\gamma_I\), Corollary~\ref{cor:sufficient-margin-condition} gives a positive recovery margin.
This is the intended regime: semantic strategies win not merely because they are informative, but because their factorization is cheaper to realize.
\end{examplebox}

\begin{examplebox}[Failure from a superficial high-information attribute]
If \(\lambda=0\) and \(\calR,\calG\) are expressive enough that \(A_{\beta,\lambda}(q)=0\) for all candidate deterministic encoders, then the profiled score is just \(I_q(Y;Z\mid X)\).
A length bucket, formatting marker, or balanced nuisance partition with larger conditional information than \(S^\star\) will beat the semantic strategy.
The theorem then correctly predicts non-recovery.
\end{examplebox}

\begin{examplebox}[Failure of strong recovery under free local relabeling]
If the available router-generator class and complexity proxy are invariant not only to global relabeling but also to input-dependent relabeling, then the encoders \(q^\pi\) in Proposition~\ref{prop:input-dependent-relabeling} can tie \(q^\star\) in profiled score.
Weak recovery may still hold, but the strong margin cannot be positive unless the objective assigns positive cost to such local label misalignment.
\end{examplebox}

\subsection{Marginal Fidelity and the Small-Complexity Limit}
\label{app:marginal-fidelity}

For a router-generator pair, define the marginal conditional law
\[
  m_{r,g}(y\mid x) := \sum_z r(z\mid x)g(y\mid x,z).
\]
The next proposition connects the profiled variational gaps to the observable marginal fit of this router-generator pair.
It clarifies how much fidelity can be lost when the profiled optimizer trades exact factorization for lower complexity.

\begin{propositionbox}[Variational gaps control marginal fidelity]
\label{prop:marginal-fidelity}
For every \((q,r,g)\),
\[
  \Expect_X \mathrm{KL}\!\left( p(\cdot\mid X) \middle\| m_{r,g}(\cdot\mid X) \right) \le \PriorGap(q,r)+\DecGap(q,g).
\]
\end{propositionbox}

\begin{proof}
Fix \(x\in\calX\). Define two joint distributions on \(\calY\times\calZ\):
\[
  P_x(y,z):=p(y\mid x)q(z\mid x,y),
  \qquad
  M_x(y,z):=r(z\mid x)g(y\mid x,z).
\]
The distribution \(P_x\) is the joint law over \((Y,Z)\) induced by first sampling
\(Y\sim p(\cdot\mid x)\) and then \(Z\sim q(\cdot\mid x,Y)\).
By Bayes' rule, it can also be written as
\[
  P_x(y,z)=q(z\mid x)p_q(y\mid x,z),
\]
with the usual convention that zero-mass branches do not affect the KL terms.

Now compare the \(Y\)-marginals of these two joint distributions. For \(P_x\),
\[
  \sum_z P_x(y,z)
  =
  \sum_z p(y\mid x)q(z\mid x,y)
  =
  p(y\mid x).
\]
For \(M_x\),
\[
  \sum_z M_x(y,z)
  =
  \sum_z r(z\mid x)g(y\mid x,z)
  =
  m_{r,g}(y\mid x).
\]
Thus \(p(\cdot\mid x)\) and \(m_{r,g}(\cdot\mid x)\) are obtained from
\(P_x\) and \(M_x\), respectively, by applying the same projection map
\((y,z)\mapsto y\).

By data processing for KL divergence, marginalizing out \(Z\) cannot increase KL.
Therefore
\[
  \mathrm{KL}\!\left(p(\cdot\mid x)\middle\|m_{r,g}(\cdot\mid x)\right)
  \le
  \mathrm{KL}(P_x\|M_x).
\]

It remains to expand the joint KL on the right. Using the factorizations
\(P_x(y,z)=q(z\mid x)p_q(y\mid x,z)\) and
\(M_x(y,z)=r(z\mid x)g(y\mid x,z)\),
\[
\begin{aligned}
  \mathrm{KL}(P_x\|M_x)
  &=
  \sum_{z,y} q(z\mid x)p_q(y\mid x,z)
  \log
  \frac{q(z\mid x)p_q(y\mid x,z)}
       {r(z\mid x)g(y\mid x,z)}
  \\
  &=
  \sum_z q(z\mid x)\log\frac{q(z\mid x)}{r(z\mid x)}
  +
  \sum_z q(z\mid x)
  \sum_y p_q(y\mid x,z)
  \log\frac{p_q(y\mid x,z)}{g(y\mid x,z)}
  \\
  &=
  \mathrm{KL}(q(\cdot\mid x)\|r(\cdot\mid x))
  +
  \Expect_{Z\sim q(\cdot\mid x)}
  \mathrm{KL}(p_q(\cdot\mid x,Z)\|g(\cdot\mid x,Z)).
\end{aligned}
\]
Combining the previous two displays gives, for each fixed \(x\),
\[
  \mathrm{KL}\!\left(p(\cdot\mid x)\middle\|m_{r,g}(\cdot\mid x)\right)
  \le
  \mathrm{KL}(q(\cdot\mid x)\|r(\cdot\mid x))
  +
  \Expect_{Z\sim q(\cdot\mid x)}
  \mathrm{KL}(p_q(\cdot\mid x,Z)\|g(\cdot\mid x,Z)).
\]
Finally, average both sides over \(X\sim\mu\). The two terms on the right become
\(\PriorGap(q,r)\) and \(\DecGap(q,g)\), respectively, which proves
\[
  \Expect_X \mathrm{KL}\!\left(
    p(\cdot\mid X) \middle\| m_{r,g}(\cdot\mid X)
  \right)
  \le
  \PriorGap(q,r)+\DecGap(q,g).
\]
\end{proof}

This proposition makes the fidelity tradeoff explicit.
For finite \(\lambda\), the profiled optimizer can prefer a simpler \((r,g)\) with nonzero variational gaps, and therefore need not exactly factorize \(p(y\mid x)\).
The next result records the high-fidelity limit.

\begin{propositionbox}[Small-\(\lambda\) fidelity for a fixed encoder]
\label{prop:lambda-zero-fidelity}
Fix \(q\in\calQ\) and \(\beta>0\).
Suppose the exact induced pair
\[
  r_q(z\mid x):=q(z\mid x), \qquad g_q(y\mid x,z):=p_q(y\mid x,z)
\]
belongs to \(\calR\times\calG\), and \(C(r_q,g_q)<\infty\).
If \((r_\lambda,g_\lambda)\) minimizes \(A_{\beta,\lambda}(q)\), for each \(\lambda>0\), then
\[
  \DecGap(q,g_\lambda)\to0, \qquad \PriorGap(q,r_\lambda)\to0
\]
as \(\lambda\downarrow0\).
Consequently,
\[
  \Expect_X \mathrm{KL}\!\left( p(\cdot\mid X) \middle\| m_{r_\lambda,g_\lambda}(\cdot\mid X) \right) \to0.
\]
\end{propositionbox}

\begin{proof}
Fix \(q\in\calQ\). By assumption, the exact induced router-generator pair
\[
  r_q(z\mid x):=q(z\mid x),
  \qquad
  g_q(y\mid x,z):=p_q(y\mid x,z)
\]
belongs to \(\calR\times\calG\) and has finite complexity.

For \(q\)'s induced router-generator pair \((r_q,g_q)\), the prior gap and decoder gap are both zero. Since \((r_\lambda,g_\lambda)\) minimizes \(A_{\beta,\lambda}(q)\), its objective value is no larger than the value achieved by the exact induced pair. Therefore
\[
\begin{aligned}
  &\DecGap(q,g_\lambda)
  + \beta \PriorGap(q,r_\lambda)
  + \lambda C(r_\lambda,g_\lambda)
  \\
  &\qquad\le
  \DecGap(q,g_q)
  + \beta \PriorGap(q,r_q)
  + \lambda C(r_q,g_q)
  =
  \lambda C(r_q,g_q).
\end{aligned}
\]
All three terms on the left are nonnegative. Hence
\[
  \DecGap(q,g_\lambda)
  \le
  \lambda C(r_q,g_q),
\]
and, since \(\beta>0\),
\[
  \PriorGap(q,r_\lambda)
  \le
  \frac{\lambda C(r_q,g_q)}{\beta}.
\]
Because \(C(r_q,g_q)<\infty\), both upper bounds go to zero as
\(\lambda\downarrow0\). Thus
\[
  \DecGap(q,g_\lambda)\to0,
  \qquad
  \PriorGap(q,r_\lambda)\to0.
\]

Finally, Proposition~\ref{prop:marginal-fidelity} gives
\[
  \Expect_X \mathrm{KL}\!\left(
    p(\cdot\mid X) \middle\| m_{r_\lambda,g_\lambda}(\cdot\mid X)
  \right)
  \le
  \PriorGap(q,r_\lambda)+\DecGap(q,g_\lambda).
\]
The right-hand side tends to zero, so the marginal KL also tends to zero.
\end{proof}

The limit \(\lambda\downarrow0\) restores high-fidelity factorization under exact realizability.
It also weakens the selection role of complexity: the profiled encoder score approaches maximum conditional information, which can favor superficial high-information partitions unless complexity still breaks the relevant near-ties.

\subsection{Empirical Recovery}
\label{app:empirical-recovery}
\label{app:finite-sample}

The population theorem gives a profiled-score margin
\(\Delta_{\beta,\lambda}^{\rho,\mathrm{prof}}(\delta)\). Empirical
recovery only requires one additional ingredient: the empirical profiled ELBO
must approximate the population profiled ELBO at a scale smaller than the same
margin measured in loss units. Since
\[
  \overline{\calL}_{\beta,\lambda}(q)
  =
  H(Y\mid X)
  -
  (1-\beta)J^{\mathrm{prof}}_{\beta,\lambda}(q),
\]
the loss separation corresponding to
\(\Delta_{\beta,\lambda}^{\rho,\mathrm{prof}}(\delta)\) is
\((1-\beta)\Delta_{\beta,\lambda}^{\rho,\mathrm{prof}}(\delta)\).
Equivalently, for every bad encoder \(q\in\calB_\rho(\delta)\),
\[
\begin{aligned}
  \overline{\calL}_{\beta,\lambda}(q)
  -
  \overline{\calL}_{\beta,\lambda}(q^\star)
  ={}&
  A_{\beta,\lambda}(q)-A_{\beta,\lambda}(q^\star) \\
  &-
  (1-\beta)
  \{I_q(Y;Z\mid X)-I_{q^\star}(Y;Z\mid X)\}.
\end{aligned}
\]
Thus empirical recovery is possible when finite samples preserve this
information-complexity loss gap.

Let \((X_i,Y_i)_{i=1}^n\) be i.i.d. samples from \(\mu(x)p(y\mid x)\). We use
the analytic expectation over \(z\sim q(\cdot\mid X_i,Y_i)\) and define
\[
\begin{aligned}
  \widehat{\calL}_{\beta,\lambda,n}(q,r,g)
  :={}&
  \frac1n\sum_{i=1}^n\sum_z q(z\mid X_i,Y_i)
  \left[
    -\log g(Y_i\mid X_i,z)
    +
    \beta\log\frac{q(z\mid X_i,Y_i)}{r(z\mid X_i)}
  \right] \\
  &\quad + \lambda C(r,g),
\end{aligned}
\]
with the usual extended-real conventions for the logarithmic terms. The
empirical profiled loss is
\[
  \widehat{\overline{\calL}}_{\beta,\lambda,n}(q)
  :=
  \inf_{r\in\calR,\;g\in\calG}
  \widehat{\calL}_{\beta,\lambda,n}(q,r,g).
\]

\begin{theorembox}[Empirical recovery for the profiled ELBO]
\label{thm:empirical-profiled-elbo-recovery}
\label{thm:empirical-uniform-convergence}
\label{cor:empirical-profiled-elbo-recovery}
Assume \(\beta<1\), \(q^\star\in\calQ\),
\(\overline{\calL}_{\beta,\lambda}(q^\star)<\infty\), and
\[
  \Delta_{\beta,\lambda}^{\rho,\mathrm{prof}}(\delta)>0.
\]
Suppose an empirical profiled minimizer exists,
\[
  \widehat q_n
  \in
  \argmin_{q\in\calQ}
  \widehat{\overline{\calL}}_{\beta,\lambda,n}(q),
\]
and the empirical profiled objective satisfies a uniform convergence bound:
\[
  \sup_{q\in\calQ}
  \left|
    \widehat{\overline{\calL}}_{\beta,\lambda,n}(q)
    -
    \overline{\calL}_{\beta,\lambda}(q)
  \right|
  <
  \frac{(1-\beta)\Delta_{\beta,\lambda}^{\rho,\mathrm{prof}}(\delta)}{2}.
\]
Then every empirical profiled minimizer recovers \(S^\star\) at scale \(\delta\):
\[
  d_\rho(\widehat q_n,S^\star)<\delta.
\]
\end{theorembox}

\begin{proof}
Let
\[
  \varepsilon_n
  :=
  \sup_{q\in\calQ}
  \left|
    \widehat{\overline{\calL}}_{\beta,\lambda,n}(q)
    -
    \overline{\calL}_{\beta,\lambda}(q)
  \right|.
\]
By empirical optimality,
\[
  \widehat{\overline{\calL}}_{\beta,\lambda,n}(\widehat q_n)
  \le
  \widehat{\overline{\calL}}_{\beta,\lambda,n}(q^\star).
\]
Therefore
\[
\begin{aligned}
  \overline{\calL}_{\beta,\lambda}(\widehat q_n)
  &\le
  \widehat{\overline{\calL}}_{\beta,\lambda,n}(\widehat q_n)
  +\varepsilon_n \\
  &\le
  \widehat{\overline{\calL}}_{\beta,\lambda,n}(q^\star)
  +\varepsilon_n \\
  &\le
  \overline{\calL}_{\beta,\lambda}(q^\star)+2\varepsilon_n \\
  &<
  \overline{\calL}_{\beta,\lambda}(q^\star)
  +(1-\beta)\Delta_{\beta,\lambda}^{\rho,\mathrm{prof}}(\delta).
\end{aligned}
\]
If \(d_\rho(\widehat q_n,S^\star)\ge\delta\), then
\(\widehat q_n\in\calB_\rho(\delta)\). Since
\[
  \overline{\calL}_{\beta,\lambda}(q)
  =
  H(Y\mid X)-(1-\beta)J^{\mathrm{prof}}_{\beta,\lambda}(q),
\]
the definition of \(\Delta_{\beta,\lambda}^{\rho,\mathrm{prof}}(\delta)\)
implies
\[
  \overline{\calL}_{\beta,\lambda}(\widehat q_n)
  \ge
  \overline{\calL}_{\beta,\lambda}(q^\star)
  +(1-\beta)\Delta_{\beta,\lambda}^{\rho,\mathrm{prof}}(\delta),
\]
a contradiction. Hence \(d_\rho(\widehat q_n,S^\star)<\delta\).
\end{proof}

This theorem is deterministic conditional on uniform convergence. The remaining
question is how large \(n\) must be for
\[
  \sup_{q\in\calQ}
  \left|
    \widehat{\overline{\calL}}_{\beta,\lambda,n}(q)
    -
    \overline{\calL}_{\beta,\lambda}(q)
  \right|
  <
  \frac{(1-\beta)\Delta_{\beta,\lambda}^{\rho,\mathrm{prof}}(\delta)}{2}
\]
to hold with high probability. This is a statistical question about the
profiled log-loss class.

In what follows, we give two concrete sufficient conditions for this uniform convergence.
The first corollary assumes a finite encoder class and a fixed concentration bound for each encoder.
The second derives a concrete condition under which such concentration holds by assuming a support floor for the decoder and router probabilities such that the log losses are bounded.

\begin{corollarybox}[Finite-class sample complexity]
\label{cor:finite-class-sample-complexity}
Suppose \(|\calQ|=N<\infty\) and for every fixed \(q\in\calQ\), the empirical
profiled loss \(\widehat{\overline{\calL}}_{\beta,\lambda,n}(q)\) satisfies
\[
  \Prob\left(
  \left|
    \widehat{\overline{\calL}}_{\beta,\lambda,n}(q)
    -
    \overline{\calL}_{\beta,\lambda}(q)
  \right|
  \ge t
  \right)
  \le
  2\exp\left(-\frac{nt^2}{2\sigma^2}\right).
\]
If
\[
  n
  \ge
  \frac{8\sigma^2}
  {(1-\beta)^2
  (\Delta_{\beta,\lambda}^{\rho,\mathrm{prof}}(\delta))^2}
  \log\frac{2N}{\eta},
\]
then with probability at least \(1-\eta\), every empirical profiled-ELBO
minimizer satisfies \(d_\rho(\widehat q_n,S^\star)<\delta\).
\end{corollarybox}

\begin{proof}
For each fixed \(q\in\calQ\), apply the assumed concentration inequality.
A union bound over the \(N\) encoders gives
\[
  \Prob\left(
  \sup_{q\in\calQ}
  \left|
    \widehat{\overline{\calL}}_{\beta,\lambda,n}(q)
    -
    \overline{\calL}_{\beta,\lambda}(q)
  \right|
  \ge t
  \right)
  \le
  2N\exp\left(-\frac{nt^2}{2\sigma^2}\right).
\]
Set
\[
  t=
  \frac{(1-\beta)\Delta_{\beta,\lambda}^{\rho,\mathrm{prof}}(\delta)}{2}.
\]
The displayed lower bound on \(n\) makes the right-hand side at most \(\eta\).
On the complementary event,
\[
  \sup_{q\in\calQ}
  \left|
    \widehat{\overline{\calL}}_{\beta,\lambda,n}(q)
    -
    \overline{\calL}_{\beta,\lambda}(q)
  \right|
  <
  \frac{(1-\beta)\Delta_{\beta,\lambda}^{\rho,\mathrm{prof}}(\delta)}{2}.
\]
The empirical recovery theorem then implies
\(d_\rho(\widehat q_n,S^\star)<\delta\) for every empirical profiled-ELBO
minimizer.
\end{proof}

Finite \(\calQ\) alone does not make the preceding concentration automatic. The
log terms
\[
  -\log g(y\mid x,z),
  \qquad
  \log\frac{q(z\mid x,y)}{r(z\mid x)}
\]
can be unbounded or infinite when \(g(y\mid x,z)\) or \(r(z\mid x)\) is near
zero. A simple sufficient condition is to assume finite
\(\calQ,\calR,\calG\), finite \(C(r,g)\), and a support floor
\(\tau\in(0,1]\) such that
\[
  r(z\mid x)\ge\tau,
  \qquad
  g(y\mid x,z)\ge\tau
\]
for all admissible \(r,g\) and all \(x,y,z\).

This is stated in the following corollary.

\begin{corollarybox}[Finite classes with bounded log loss]
\label{cor:finite-class-bounded-profiled-support-floor}
Suppose \(\calQ,\calR,\calG\) are finite, \(C(r,g)<\infty\) for all
\((r,g)\in\calR\times\calG\), and there is \(\tau\in(0,1]\) such that
\[
  r(z\mid x)\ge\tau,
  \qquad
  g(y\mid x,z)\ge\tau
\]
for all \(r\in\calR\), \(g\in\calG\), and all \(x,y,z\).
If
\[
  n
  \ge
  \frac{2(1+\beta)^2\log^2(1/\tau)}
  {(1-\beta)^2
  (\Delta_{\beta,\lambda}^{\rho,\mathrm{prof}}(\delta))^2}
  \log\frac{2|\calQ||\calR||\calG|}{\eta},
\]
then with probability at least \(1-\eta\), every empirical profiled-ELBO
minimizer satisfies \(d_\rho(\widehat q_n,S^\star)<\delta\).
\end{corollarybox}

\begin{proof}
For a fixed triple \((q,r,g)\), the data-dependent part of the single-sample
loss is
\[
  \sum_z q(z\mid x,y)
  \left[
    -\log g(y\mid x,z)
    +
    \beta\log\frac{q(z\mid x,y)}{r(z\mid x)}
  \right].
\]
The support floor gives \(-\log g(y\mid x,z)\le\log(1/\tau)\), and
\[
  \sum_z q(z\mid x,y)\log\frac{q(z\mid x,y)}{r(z\mid x)}
  =
  \mathrm{KL}(q(\cdot\mid x,y)\|r(\cdot\mid x))
  \le
  \log\frac1\tau.
\]
Thus the data-dependent part is bounded between \(0\) and
\((1+\beta)\log(1/\tau)\). The complexity term \(\lambda C(r,g)\) is
deterministic and cancels in empirical-population deviations.

Hoeffding's inequality and a union bound over
\(\calQ\times\calR\times\calG\) imply
\[
  \Prob\left(
  \sup_{q,r,g}
  \left|
    \widehat{\calL}_{\beta,\lambda,n}(q,r,g)
    -
    \calL_{\beta,\lambda}(q,r,g)
  \right|
  \ge t
  \right)
  \le
  2|\calQ||\calR||\calG|
  \exp\left(-\frac{2nt^2}{(1+\beta)^2\log^2(1/\tau)}\right).
\]
Set
\[
  t=
  \frac{(1-\beta)\Delta_{\beta,\lambda}^{\rho,\mathrm{prof}}(\delta)}{2}.
\]
The displayed lower bound on \(n\) makes the right-hand side at most \(\eta\).

On the complementary event, profiling cannot increase the deviation: for every
\(q\),
\[
  \left|
    \inf_{r,g}\widehat{\calL}_{\beta,\lambda,n}(q,r,g)
    -
    \inf_{r,g}\calL_{\beta,\lambda}(q,r,g)
  \right|
  \le
  \sup_{r,g}
  \left|
    \widehat{\calL}_{\beta,\lambda,n}(q,r,g)
    -
    \calL_{\beta,\lambda}(q,r,g)
  \right|.
\]
Therefore
\[
  \sup_{q\in\calQ}
  \left|
    \widehat{\overline{\calL}}_{\beta,\lambda,n}(q)
    -
    \overline{\calL}_{\beta,\lambda}(q)
  \right|
  <
  \frac{(1-\beta)\Delta_{\beta,\lambda}^{\rho,\mathrm{prof}}(\delta)}{2}.
\]
The empirical recovery theorem then gives
\(d_\rho(\widehat q_n,S^\star)<\delta\) for every empirical profiled-ELBO
minimizer.
\end{proof}

Support floors are only one transparent sufficient condition. Other standard
routes are to clip predictive probabilities, assume direct concentration of the
profiled losses, prove a high-probability envelope or sub-exponential tail bound
for the relevant log losses, or use entropy, bracketing, or Rademacher control
for the likelihood class. The recovery argument itself only needs uniform
accuracy at scale
\((1-\beta)\Delta_{\beta,\lambda}^{\rho,\mathrm{prof}}(\delta)\); support or
tail assumptions are one way to obtain that event.

\end{document}